\documentclass[11pt]{article}
\usepackage{xr}
\usepackage{tikz}
\usetikzlibrary{positioning}
  \usepackage{pgf}
  \usepackage{colortbl}
\usepackage{amsmath,amssymb}
\usepackage{float}
\usepackage{bbm}
\usepackage{xcolor}
\usepackage{enumerate}
\usepackage[authoryear]{natbib}
\usetikzlibrary{arrows,positioning,automata,calc,fit,shapes.geometric,arrows.meta}
\usepackage{geometry}   

\usepackage{rotating,multirow,makecell,array}
\usepackage{caption}
\usepackage{subcaption}
 \usepackage{bbm}

\usepackage{subfiles}
\usepackage{scalerel,stackengine}
\usepackage{lipsum}
\stackMath
\newcommand\reallywidehat[1]{%
\savestack{\tmpbox}{\stretchto{%
  \scaleto{%
    \scalerel*[\widthof{\ensuremath{#1}}]{\kern-.6pt\bigwedge\kern-.6pt}%
    {\rule[-\textheight/2]{1ex}{\textheight}}
  }{\textheight}%
}{0.5ex}}%
\stackon[1pt]{#1}{\tmpbox}%
}
\usepackage[english]{babel}
\usepackage{amsthm}
\usepackage{bbm}
\usepackage{blindtext}
\usepackage{dsfont}
\usepackage{algorithm}
\usepackage{algpseudocode}
\usepackage{amsmath,amssymb}
\usepackage{multirow,makecell}

\usepackage[utf8x]{inputenc}

\makeatletter
\def\BState{\State\hskip-\ALG@thistlm}
\makeatother
\usepackage{amssymb}
\usepackage[inline]{enumitem}
\usepackage{url} 

\DeclareMathOperator*{\argmin}{arg\,min}

\newcommand{\norma}[1]{\left\lVert#1\right\rVert}

\usepackage[pdftex,colorlinks]{hyperref}

  \hypersetup{
colorlinks =true,
citecolor    = blue,
citebordercolor = violet,
filebordercolor=red,
linkbordercolor=blue
}
\usepackage{authblk}
\usetikzlibrary{shapes.geometric}

\usetikzlibrary{shapes,snakes}
\makeatletter
\newcommand*{\addFileDependency}[1]{
  \typeout{(#1)}
  \@addtofilelist{#1}
  \IfFileExists{#1}{}{\typeout{No file #1.}}
}
\makeatother
 
\definecolor{coquelicot}{rgb}{0.20, 0.12, 0.72}

 \providecommand{\keywords}[1]{\noindent \textbf{\textbf{Keywords:}} #1}

\def\spacingset#1{\renewcommand{\baselinestretch}%
{#1}\small\normalsize} \spacingset{1}

\title{\sc{\LARGE DeepHazard: \\ \LARGE neural network for time-varying risks\thanks{ We thank Alex Cloninger, Ronghui Xu  and participants of Stanford Biostatistics Workshop on helpful comments and discussions. Authors can be reached at   {\href{mailto:drava@ucsd.edu}{drava@ucsd.edu} and \href{mailto:jbradic@ucsd.edu}{jbradic@ucsd.edu}, respectively.}}}}
\date{\today}

\author[1]{Denise Rava}
\author[1,2]{Jelena Bradic}
\affil[1]{Department of Mathematics, University of California, San Diego}
\affil[2]{Halicio\u{g}l{u} Data Science Institute, University of California, San Diego}

\begin{document}

\maketitle

\begin{abstract}
Prognostic models in survival analysis are  aimed at understanding the relationship between patients' covariates and the distribution of survival time. Traditionally, semi-parametric models, such as the Cox model, have been assumed. These often rely on strong  proportionality assumptions of the hazard that might be violated in practice. Moreover, they  do not often include covariate information updated over time.  We propose a new flexible method for survival prediction: DeepHazard, a neural network for time-varying risks. Our approach is tailored for a wide range of continuous hazards forms, with the only restriction of being additive in time. A flexible implementation, allowing  different optimization methods, along with any norm penalty, is developed. Numerical examples illustrate  that our approach outperforms existing state-of-the-art methodology in terms of predictive capability evaluated through the C-index metric. The same is revealed on the popular real datasets as METABRIC, GBSG, and ACTG.

\smallskip
\end{abstract}
\newpage
\section{Introduction}
Understanding the relationship between covariates and the distribution of survival time is fundamental in many fields spanning medicine, biology, healthcare, economics, and engineering. Survival data are often incomplete due to censoring, making the traditional predictive methods unsuitable. Traditionally, several semiparametric survival models, such as the popular Cox Model \citep{cox1972regression}, the Additive Hazards Model \citep{aalen1980model} or the Accelerated Failure Time model \citep{wei1992accelerated}, have been proposed and extensively used.
Developed to deal with censoring; however, they model the hazards as a particular function of a linear combination of the data, limiting their applicability in many real-world applications.

To overcome this difficulty, the interest in using deep learning methods, such as neural networks, for survival prediction has been increasing. 
Several nonparametric extensions of the Cox Model have appeared in the literature; see, for example, \cite{faraggi1995neural,ching2018cox,liao2016combining,zhu2016deep,katzman2018deepsurv,kvamme2019time}. They make use, to train the neural network, of the classical Cox partial likelihood and base their analysis on the proportional hazard assumption. The latter is often unrealistic and represents a relevant limitation. Non-proportional hazards are widely occurrent: when the effect of a treatment vanishes over time, and henceforth the ratio of the hazards tends to one, or when a drug is beneficial for one subgroup but harmful for the other, resulting in crossing survival curves. Non-proportional hazards are difficult to model. They usually indeed don't allow the use of a flexible and nonparametric baseline hazard.

Another line of work pertains the usage of discrete-time hazards for survival prediction; see for example, \cite{liestbl1994survival,brown1997use,biganzoli1998feed,zhu2016deep,luck2017deep,fotso2018deep,lee2018deephit,gensheimer2019scalable,grisandeep,ren2019deep,zhao2019dnnsurv,lee2019dynamic}. 
They don't make assumptions on the form of the hazard; however, they treat survival time as a discrete random variable taking only finitely many pre-determined values, loosing, therefore, the continuous nature of the problem itself. Moreover, they often cast the survival problem as a classification one, considering every observation as a sequence of zeros and ones to indicate their status. Naturally, with discrete approaches, the hazard is no longer a rate but a conditional probability.
A different approach is the one proposed by \cite{zhao2019dnnsurv}. The authors reduce the survival problem to a standard regression problem by considering inputting the missing outcomes with Kaplan Meier survival estimates. However, regression on such pseudo-responses is deemed biased whenever data is not missing at random.
We construct, instead, a new survival neural network.

To overcome these limitations, we propose  DeepHazard, a new neural network that doesn't rely on the assumption of proportional hazards while not neglecting the continuous nature of the data. Our approach is indeed tailored for a wide range of hazards, with the only restriction of being continuous and additive in time. Illustrative examples include a case where the effect of treatment or the treatment status changes with time;  some patients are treated only after their disease progresses.

Building on the promising alternative of the Cox model, the non-parametric additive hazards model, we propose a new non-parametric alternative of the additive hazards loss. The latter doesn't constrain the risk of being of a particular form or being constant in time. Moreover, it naturally incorporates time-dependent covariates making our approach suitable for a large class of real data applications.
In particular, our approach is designed to treat an aligned type of data arising whenever for each observation, and each covariate, a sequence of measurement at different time points is available, for example, in a series of follow-up visits. 

The sequential nature of the data is incorporated by dividing the data in multiple time-frames and building a neural network in each time-frame to estimate the time-varying risk. Each neural network is trained on the observations, still at risk. Moreover, the interdependency between different time-frames is directly assimilated by adding to the input of every time interval-specific neural network, the output of the network built in the previous time-period.
Figure \ref{fig:1} presents one possible architecture. The output node (blue in Figure \ref{fig:1}) of each of the time-frames, denotes the predicted value of the risk score at that period. The input nodes (red in Figure \ref{fig:1}) in each of the time-frames denote at-risk observations at that time-frame. Note that they change both in numbers and type from time-frame to time-frame.
For the proposed neural network, the steps of feature extraction and survival analysis are not separated or done through two separate optimization procedures. They are gathered in one unique neural network, and the optimization of all the parameters happens together using the proposed survival loss. In this way, observations still at-risk are kept together.

DeepHazard outputs, for each combination of covariates, a rich estimate of the survival function, including the baseline survival, as well as survival in desired time-intervals, therefore allowing a deep understanding of the time to event distribution and comparison between different groups and observations. The performance of our approach is evaluated through extensive simulations. We show that our method outperforms existing methods in terms of predictive capability, evaluated through the time-dependent C-index metric \citep{antolini2005time}. 
We also apply DeepHazard to the popular real datasets: METABRIC, GBSG, and ACTG to study time to death of breast-cancer and HIV-infected patients.

\subsection{Related literature}
Different methods that make use of machine learning techniques have been employed to analyze continuous survival data.
Random survival forest of \citet{ishwaran2008random} extends the random forest methodology to survival analysis. Recently broadened to accommodate time-varying covariates, \citep{wongvibulsin2020clinical}, random survival forest consists of an ensemble of survival trees that are grown following a particular splitting rule that aims to maximize the difference between estimated survival curves in children nodes. Although a model is not explicitly assumed, the random survival forest's predictive performance depends on the splitting rule chosen. The most popular uses log-rank split statistics, which is known to lack power when the proportional hazards assumption is violated.

Machine learning techniques for discrete-time survival data include DeepHit and Dynamic-DeepHit, \citep{lee2018deephit,lee2019dynamic}, a neural network that directly estimates the probability mass function of experiencing a particular event at a specific time.
\citet{fotso2018deep} recasts the output of observation as a sequence of zeros (up to the event time) followed by a sequence of ones (after the event time) and applies the framework of neural networks to the multi-task logistic regression. 
\citet{kvamme2019continuous} rewrites the output as a vector of zeros with a single one corresponding to the observed event and makes use of the negative log-likelihood for Bernoulli data to train the neural network. The authors then propose an extension to continuous-time survival data using discretization and interpolation strategies.
\citet{zhong2019survival} introduce the stacking idea that recasts the data into a large data frame where the output column is a series of zeros and ones. The problem is then treated as a classification problem onto which various existing techniques can be directly applied.

When the time is not discretized and is treated as continuous, semi-parametric approaches based on the popular Cox model have been proposed.
\citet{katzman2018deepsurv} parametrizes a Cox regression model with a neural network building on the work of \citet{faraggi1995neural}.
\citet{kvamme2019time} proposes an extension of it introducing an approximation of the partial log-likelihood to batches of data and allowing the relative risk function to depend on time. In both cases, the model is a relative risk model that does not allow the introduction of time-dependent covariates.
A fully parametric approach has recently been proposed by \citet{nagpal2020deep}, where the survival function conditional on the fixed (not time-dependent) covariates is assumed to be a mixture of individual parametric survival distributions.

In this work, we build on the literature of semiparametric models for continuous-time survival data, proposing a different loss function, entirely unrelated to the partial likelihood typical of the proportional hazards model. Moreover, we propose a framework that allows the extension of our and potentially many other neural network methodologies to time-dependent covariates.

\subsection{Organization of the paper}

Section \ref{lab:dh} contains the details of the proposed DeepHazard algorithm which includes a new time-additive hazards model, Section \ref{lab:model}, a decomposition of the loss function, Section \ref{lab:loss}, as well as the details of the estimation and prediction, Section \ref{lab:estimation} and \ref{lab:prediction}, respectively. Section \ref{lab:3} includes detailed finite sample experiments on time-dependent covariates and outcomes where we illustrate the impact of censoring, sample size, time, and feature space. Section \ref{lab:4} focuses on real data examples where we compare with the Random Survival Forest and DeepSurv algorithms and demonstrate superior performance.

\def\layersep{1.4cm}
 \definecolor{amber}{rgb}{1.0, 0.75, 0.0}
\definecolor{bazaar}{rgb}{0.6, 0.47, 0.48}
\definecolor{babyblueeyes}{rgb}{0.63, 0.79, 0.95}
\definecolor{bleudefrance}{rgb}{0.19, 0.55, 0.91}
\definecolor{britishracinggreen}{rgb}{0.0, 0.26, 0.15}
\definecolor{cardinal}{rgb}{0.77, 0.12, 0.23}
\definecolor{charcoal}{rgb}{0.21, 0.27, 0.31}

\begin{figure}
\centering
\begin{tikzpicture}[shorten >=1pt,->,draw=black!40, node distance=\layersep]
    \tikzstyle{every pin edge}=[<-,shorten <=1pt]
    \tikzstyle{neuron}=[circle,fill=black!25,minimum size=15pt,inner sep=0pt]
    \tikzstyle{input neuron}=[neuron, fill=cardinal!80];
    \tikzstyle{output neuron}=[neuron, fill=bleudefrance];
    \tikzstyle{hidden neuron}=[neuron, fill=charcoal!80];
    \tikzstyle{annot} = [text width=4em, text centered]

    \foreach \name / \y in {1,...,6} 
        \node[input neuron, pin=left:$D_0$] (I-\name) at (0,-\y) {};
        
          \foreach \name / \y in {9,10,11,12,13} 
        \node[input neuron, pin=left:$D_1$ ] (I1-\name) at (\layersep + \layersep+\layersep+0.3cm ,-\y cm){};
        
         \foreach \name / \y in {3,4,5,6} 
        \node[input neuron, pin=left:$D_2$ ] (I2-\name) at (\layersep + \layersep+\layersep+\layersep+\layersep + \layersep+\layersep -0.5cm ,-\y cm){};

    
    \foreach \name / \y in {1,...,8}
        \path[yshift=1.0cm]
            node[hidden neuron] (H-\name) at (\layersep,-\y cm) {};

      \foreach \name / \y in {8,...,15}
        \path[yshift=1.0cm]
            node[hidden neuron] (H2-\name) at (\layersep + \layersep+\layersep  + \layersep +1cm  ,-\y cm) {};

     \foreach \name / \y in {1,...,8}
        \path[yshift=1.0cm]
            node[hidden neuron] (H4-\name) at (\layersep + \layersep+ \layersep+\layersep  + \layersep + \layersep+\layersep  + \layersep ,-\y cm) {};
            
 \foreach \name / \y in {2,4,6}
        \path[yshift=1.0cm]
            node[hidden neuron] (H1-\name) at (\layersep + \layersep  ,-\y cm) {};

                \foreach \name / \y in {8,9,13,15}
        \path[yshift=1.0cm]
            node[hidden neuron] (H3-\name) at (\layersep + \layersep+\layersep  + \layersep +\layersep+1cm  ,-\y cm) {};

     \foreach \name / \y in {1,4,7}
        \path[yshift=1.0cm]
            node[hidden neuron] (H5-\name) at (\layersep + \layersep+ \layersep+\layersep  + \layersep +\layersep  + \layersep +\layersep  + \layersep ,-\y cm) {};

    \node[output neuron,  right of=H1-2] (O) {};
    

  \node[output neuron,right of=H3-9] (O1) {};
  
  \node[output neuron,pin={[pin edge={->}]},right of=H5-4] (O2) {};
  
    \foreach \source in {1,...,6}
        \foreach \dest in {1,...,8}
            \path (I-\source) edge (H-\dest);
            
                \foreach \source in {1,...,8}
        \foreach \dest in {2,4,6}
            \path (H-\source) edge (H1-\dest);

    \foreach \source in {2,4,6}
        \path (H1-\source) edge (O);
        
              \foreach \source in {1,...,6}
        \foreach \dest in {8,...,15}
            \path (O) edge (H2-\dest);
            
               \foreach \source in {9,10,11,12,13}
        \foreach \dest in {8,...,15}
            \path (I1-\source)  edge (H2-\dest);
            
                     \foreach \source in {8,...,15}
        \foreach \dest in {8,9,13,15}
            \path (H2-\source) edge (H3-\dest);
            
                \foreach \source in {8,9,13,15}
        \path (H3-\source) edge (O1);
        
     \foreach \source in {3,4,5,6}
        \foreach \dest in {1,...,8}
            \path (I2-\source)  edge (H4-\dest);

        \foreach \dest in {1,...,8}
            \path (O)  edge (H4-\dest);
            
             \foreach \dest in {1,...,8}
            \path (O1)  edge (H4-\dest);
            
              \foreach \source in {1,...,8}
        \foreach \dest in {1,4,7}
            \path (H4-\source) edge (H5-\dest);
            
               \foreach \source in {1,4,7}
        \path (H5-\source) edge (O2);
            
    \node[annot,above of=H-1, node distance=1cm] (hl) {Hidden layer};
     \node[annot,above of=H1-2, node distance=2cm] (hl1) {Dropout layer};
          \node[annot,above of=H2-8, node distance=8cm] (hl2) {Hidden layer};
          \node[annot,above of=H3-8, node distance=8cm] (hl3) {Dropout layer};
            \node[annot,above of=H4-1, node distance=1cm] (hl4) {Hidden layer};
                \node[annot,above of=H5-1, node distance=1cm] (hl5) {Dropout layer};
    \node[annot,left of=hl] {Input layer};
    \node[annot,right of=hl1] {Time 1};
      \node[annot,right of=hl3] {Time 2};
       \node[annot,right of=hl5] {Time 3};
       
            \node[annot,above of=O, node distance=0.5cm]  {$h_0$} ;
             \node[annot,above of=O1, node distance=0.5cm]  {$h_1$} ;
              \node[annot,above of=O2, node distance=0.5cm]  {$h_2$} ;
\end{tikzpicture}
\caption{Example of DeepHazard architecture:  The output  node (blue) of  each time-frame, denotes  the predicted value of the hazard at that time period.  The input nodes (red) in each of the time-frames denote at-risk observations at that time-frame.}\label{fig:1}
\end{figure}
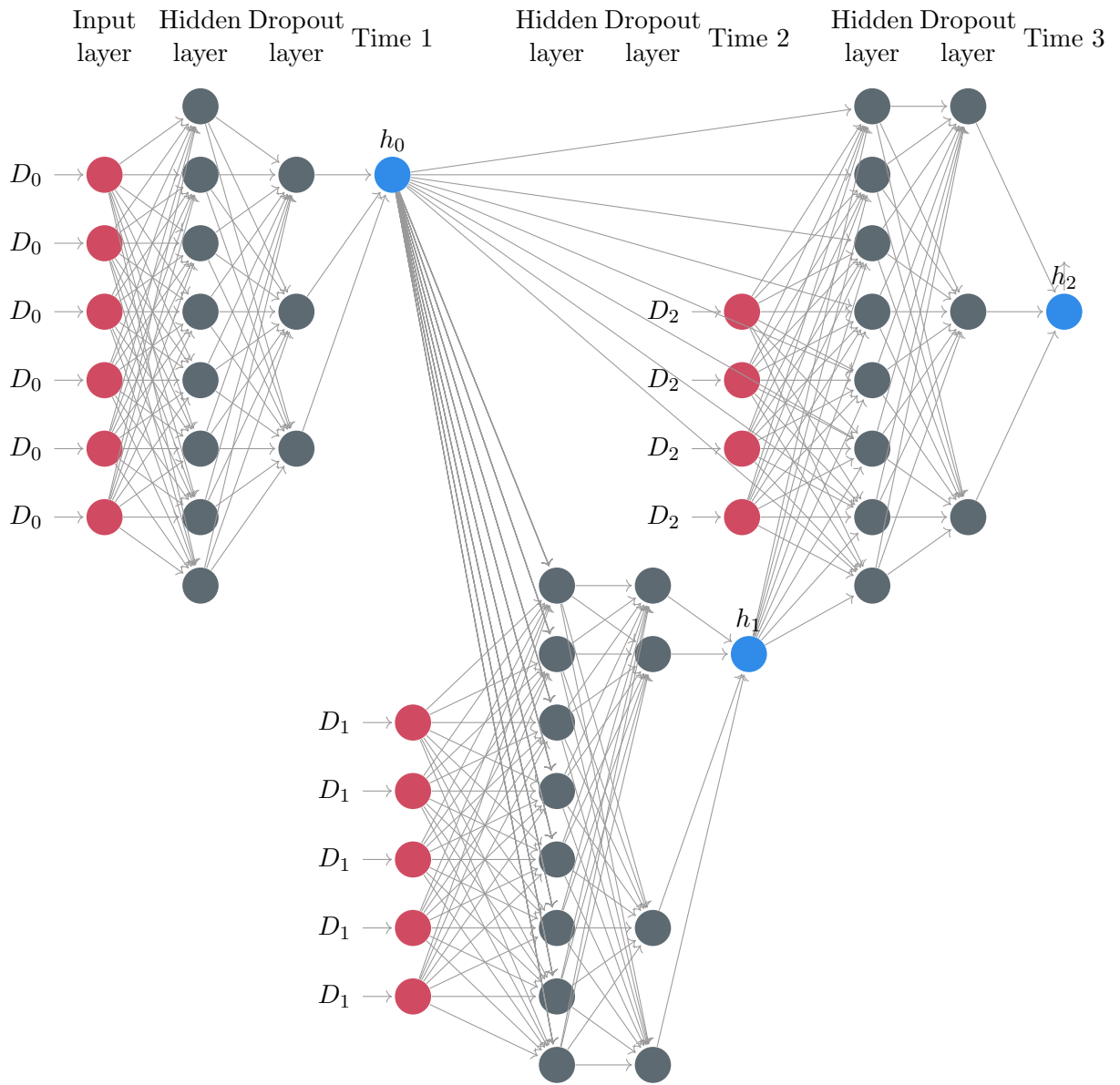

\section{DeepHazard learning} \label{lab:dh}
We introduce a new survival model, additive in time only, that explains the survival of a subject given, possibly time-varying, covariates.

Observations of survival times are often censored. This is the case when a patient drops out of a hospital or drug-treatment study. The time of death is, in this case, never observed; however, we know that the patient was still alive when he left the study. This is modeled with a random variable $C$. If $T$ denotes the survival time, then the censored observations regarding the outcome of interest are denoted with $X = \min\{T,C\}$. Together with $X$ we typically assume that an event indicator, $\delta$ is observed; here, $\delta=\mathbbm{1}\{T\leq C\}$. 

Medical studies are typically monitored in regular time intervals where a set of personal, medical information is collected, such as blood pressure, drugs taken, temperature reading, oxygenation of the blood. Some of those can naturally be treated as baseline variables, i.e., variables not changing with time; examples include gene expressions of particular tumor tissue, demographics, age. However, the majority are time-varying. For simplicity in notation, we denote all of the covariates as time-varying variables $Z(t) \in \mathbb{R}^p $.

\subsection{Time additive hazards model} \label{lab:model}
We propose a new model, additive in time, that assumes that the hazard function, 
$$\lambda(t \mid Z(t)) = \lim _{h\to 0} \frac{ P\left(T\in(t,t+h] \bigl|T\geq t, Z(t)\right)}{h},$$ is the sum of two components, a baseline hazard $\lambda_0(t)$ that depends only on time and a risk score, $h(Z(t),t)$ that encloses the effect of the individual's covariates $Z(t)$, possibly time-varying, onto the hazard. The hazard is interpreted in the standard way, as  the probability of an event in the interval $[t,t+dt)$ given covariate $Z(t)$ and assuming that no previous event has happened.

We assume that the covariates are measured at a sequence of $M$ time points (follow up visits),
$$t_0,t_1,\ldots,t_M.$$ Let's notice that we don't require $t_0,t_1,\ldots,t_M$ to be the same as event times.
Therefore, we naturally divide the time into a sequence of intervals $[t_0,t_{1}),\ldots,[t_M,\infty)$.
For example, let us assume that every patient is subjected to a visit every two months, and at every such visit, a series of physical values, as blood pressure, is measured and recorded. In this case, we would have as intervals $[0,2),[2,4),\ldots$, and the series of the measured values will be encoded as $Z(0), Z(2),\dots$. 
We assume that at any intervals $[t_j,t_{j+1})$ the risk score of a subject is described by a constant in time risk score $h_j$. 
\[
h(Z(t),t)= h_j(Z(t)) , \qquad t\in [t_j,t_{j+1}), \qquad j=0,1,\dots, M,
\]
To acknowledge the continuous nature of the time and the natural possible dependence   onto the past values, we allow the risk score $h_j$ to depend on previous-in-time risk scores $h_0,\ldots,h_{j-1}$.
In other words $h(Z(t),t)$ satisfies
\begin{equation}\label{hstruct}
 h_j(Z(t))= f_j(Z(t_j), h_0(Z(t)), h_1(Z(t)), \cdots, h_{j-1}(Z(t))), \quad t\in [t_j,t_{j+1}), \quad j=0,1,\dots, M,
\end{equation}
where $f_j$ is an unknown function.
This describes a recursive relationship
\begin{align*}
h_0(Z(t))&= {f}_0(Z(t_0)), \  h_1(Z(t)) = {f}_1 \bigl( Z(t_1),  {f}_0( Z(t_0)) \bigl), \  
\\
h_2(Z(t))&= {f}_2\Bigl( Z(t_2),  {f}_0(Z(t_0)) ,  {f}_1 \Bigl( Z(t_1), {f}_0(Z(t_0)) \Bigl) \Bigl), \cdots.
\end{align*}
With a small abuse in notation we   drop the notation $f_j$ and use $h_j$ to denote the unknown functional relationship at time interval $j$.

Therefore, primarily  we consider  the following representation of the hazard
\begin{equation}\label{model}
\lambda(t \mid Z(t))=\lambda_0(t)+h(Z(t),t)
\end{equation}
where 
\begin{equation}\label{modelassumption}
h(Z(t),t)=\sum_{j=0}^M h_j \Bigl(Z(t_j),h_0(Z(t)),\ldots,h_{j-1}(Z(t))\Bigl)\mathbbm{1}\left(t_{j}\leq t < t_{j+1}\right),
\end{equation}
where $t_{M+1}=\infty$ and $h_0(Z(t)), \ldots, h_M(Z(t))$ are functions of the covariates.

 The form of model \eqref{model} is reminiscent of the traditional additive hazards model \citep{aalen1980model}, which takes the following form, $\lambda(t \mid Z)=\lambda_0(t)+\beta(t)Z(t)$ with the risk being limited to be of a linear form.
The proposed model extends it to comprise a broader range of risk score forms and to incorporate the sequential nature of time-varying covariates.

{\bf Example 1:}  Sum of all previous in time hazards:
\begin{align*}
h_0(Z(t)) &=  f_0(Z(t_0))   ,  \ h_1(Z(t)) = h_0(Z(t))  +f_1(Z(t_1))    , 
\\ 
h_j(Z(t))  &=  h_0(Z(t))     + \dots + h_{j-1}(Z(t))  + f_j(Z(t_j))  
\end{align*}
This can be named nonparametric additive hazards model; structure of the hazard mimics that of generalized additive models \citep{hastie1990generalized}.  Similarly, one can consider sum of the last few in time hazards only.

{\bf Example 2:}  Product of the last  $k$  hazards:
\begin{align*}
h_0(Z(t)) & =  f_0(Z(t_0)),  \ h_1(Z(t))  =  f_0(Z(t_0)) f_1(Z(t_1)), \ 
\\
h_j(Z(t)) & = h_{j-k}(Z(t))      \cdots   h_{j-1}(Z(t))    f_j(Z(t_j))
\end{align*}
Here the logarithm of the hazard has nonparametric and additive structure. However the logarithmic transformation as well as functions $f_0,\dots,f_M$ are unknown a-priori.  

{\bf Example 3:} Heterogeneous hazard:
\[
h_j(Z(t))  = \sigma_j   Z(t_j) \qquad \sigma_j^2 = \omega_j+ \alpha_j  f_{j-1}^2(Z (t_{j-1}))   + \beta_j \sigma^2_{j-1}
\]
where $\omega >0$, $\alpha, \beta \geq 0$. The aforementioned constants as well as functions $f_j$ are all unknown parameters of the hazard.   In particular,
\begin{align*}
h_0(Z(t))  &= \sigma_0 Z (t_0), \sigma_0 ^2= \omega_0 >0,\\
  h_1(Z(t))  &= \sigma_1 Z (t_1) , \sigma_1^2 = \omega_1 + \alpha_1(f_0 (Z(t_0)) - \theta_1 \omega_0)^2 + \beta \omega_0^2, \cdots
\end{align*}

More in general, it is easy to see how any survival model can be written as equation \eqref{model}. Our modeling Assumption, \eqref{modelassumption}, on the form of $h(Z(t),t)$ can be seen as an approximation for estimation purposes. Indeed, we only assume the risk to be constant into intervals. Moreover, we allow the dependency of the risk score, of a specific interval, onto the risk scores of the previous intervals, making the assumption of piecewise constant risk less strict and allowing the continuous nature of the time to play an explicit role. Therefore, our model can be applied to a wide variety of risk score forms. As long as the intervals are dense enough, and the smoothness of the risk score is adequate, our approximation will work well.   
 
 \subsection{Quadratic loss function} \label{lab:loss0}
In this section we want to motivate our score function or loss function through a population perspective first. 
In the following we use $Y(t)=\mathbbm{1}({X\geq t})$ to denote  the at-risk indicator, i.e.  subset of observations which are at time $t$ still at risk of experiencing an ``event,'' i.e., death.  In addition,  we indicate with $N(t)=\mathbbm{1}({X\leq t, \delta=1})$  the counting process of whether and when an ``event'' has occurred.

 The estimation strategy  borrows techniques from the additive hazards model and its least squares loss therefore landing itself particularly useful for  neural-network approaches.
If indeed, we consider the generic representation of the model \eqref{model},
\begin{equation}\label{martdec}
dN(t)=\lambda(t \mid Z(t))Y(t)dt+dM(t)
\end{equation}
where $M(t)$ is the associated martingale process, the following least squares loss, also called in the literature least-squares contrast, \citep{reynaud2006penalized}, for a generic function $f(Z(t),t)$, can be easily derived:
\[
\gamma(f)=-\frac{1}{n}\sum_{i=1}^n\int_0^\tau f(Z_i(t),t) dN_i(t)+\frac{1}{2n}\sum_{i=1}^n\int_0^\tau f(Z_i(t),t)^2Y_i(t)dt.
\]
In the above, $\tau$ is an upper bound of time due to administrative censoring.
Taking the expected value on both sides of  \eqref{martdec} and considering the martingale decomposition, we get:
\[
\mbox{E}\left\{\gamma(f)\right\}=-\mbox{E}\left\{\int_0^\tau f(Z_i(t),t) \lambda(t  \mid Z_i(t))Y_i(t)dt\right\}+\frac{1}{2}\mbox{E}\left\{\int_0^\tau f(Z_i(t),t)^2Y_i(t) dt\right\}
\]
Defining with $\|{\cdot}\|_\mu$, the following norm:
$
\norma{g}_\mu=\mbox{E}\left\{\int_0^\tau f^2(Z(t),t)Y(t)dt\right\}
$
we are left with
\begin{align*}
2\mbox{E}\left\{\gamma(f)\right\}&=\mbox{E}\left[\int_0^\tau \left\{f(Z(t),t)-\lambda(t \mid Z(t))\right\}^2 \lambda(t \mid Z(t))Y_i(t)dt\right]-\mbox{E}\left\{\int_0^\tau \lambda(t\mid Z(t))^2Y_i(t) dt\right\}\\
&=\norma{f(Z(t),t)-\lambda(t\mid Z(t))}_\mu-\norma{\lambda(t\mid Z(t))}_\mu.
\end{align*}
The latter justifies the minimization of the least squares contrast as estimation strategy for the hazard function $\lambda(t \mid Z(t))$, as explained in \citet{comte2011adaptive}.
If we consider our additive form of the hazard \eqref{model}, $f(Z(t),t)=\lambda_0(t)+h(Z(t),t)$, the loss can be decomposed as follows:
\[
\gamma(f)=\gamma_1(\lambda_0)+\gamma_2(h)+\gamma_3(\lambda_0,h)
\]
where 
\[
\gamma_1(\lambda_0)=\frac{1}{2n}\sum_{i=1}^n\int_0^\tau \left\{\lambda_0(t)+\Bar{h}(t)\right\}^2Y_i(t)dt-\frac{1}{n}\sum_{i=1}^n\int_0^\tau \left\{\lambda_0(t)+\Bar{h}(t)\right\}dN_i(t)
\]
\[
\gamma_2(h)=\frac{1}{2n}\sum_{i=1}^n\int_0^\tau \left\{h(Z_i(t),t)-\Bar{h}(t)\right\}^2Y_i(t)dt-\frac{1}{n}\sum_{i=1}^n\int_0^\tau \left\{h(Z_i(t),t)-\Bar{h}(t)\right\}dN_i(t)
\]
\[
\gamma_3(\lambda_0,h)=\frac{1}{n}\sum_{i=1}^n\int_0^\tau \left\{h(Z_i(t),t)-\Bar{h}(t)\right\} \left\{\lambda_0(t)+\Bar{h}(t)\right\}Y_i(t)dt
\]
where
$$\Bar{h}(t)=\frac{\sum_{i=1}^{n}h( Z_i(t),t)Y_i(t)}{\sum_{i=1}^{n}Y_i(t)}.$$
By easy computation it can be proven that $\gamma_3(\lambda_0,h)=0$. It is therefore suitable, for estimation of the risk $h(Z(t),t)$,  to consider the minimization of $\gamma_2(h)$ solely. Details of the above decompositions can be found in \citet{gaiffas2012high}.
In our approach, we make use of a regularized version of $ \gamma_2(h)$,
\[
\min _h  \left\{\gamma_2(h) + P(h) \right\}
\]
 where $P$ is an appropriate penalty function of practitioners choice. We show the details in the next section.

\subsection{Loss function decomposition} \label{lab:loss}
Noticing that  time-dependent covariates are observed in a natural, sequential ordering, $t_0 \leq t_1 \leq \dots \leq t_M$, and because of the assumed form of the hazard \eqref{modelassumption}, to estimate the risk $h(Z(t),t)$, we need to estimate the various time-intervals specific risk $h_j$ for $j=0,\ldots,M$. Intuitively, it makes sense to involve in the estimation of each $h_j$, only the observations at risk on the $j$th interval, discarding everyone that is censored or have experienced the events before the start of that particular interval. In the following, we explain the mathematical arguments in detail.

In our approach, every $h_j$ will be estimated by a neural network $j$, whose parameters, biases and weights, will be indexed by $\theta_j$.
In the following we use the generic $\theta$ to indicate the collection of $(\theta_0,\ldots,\theta_M)$ and we use $h_\theta$ to denote the dependency, explained in details later, of the final estimate of $h(Z(t),t)$ onto the parameters of the networks.
Henceforth, we make use of the following regularized version of $ \gamma_2(h)$:
\begin{equation}\label{loss1}
\gamma_2(h_\theta) + \lambda \sum_{j=0}^M\norma{\theta_j}_p,
\end{equation}
where we implemented two norms: $p = {1,2}$ to allow for both the Lasso and the Ridge penalty.

We observe that the integrals in \eqref{loss1} can be broken down as sums of $M+1$ integrals,  one for each time intervals introduced above, as in the following: 
\begin{align}\label{loss21}  
\sum_{j=0}^M \mathcal{L}_{j}(\theta_j) + \lambda \norma{\theta_j}_p
\end{align}
where
 \begin{align}\label{loss2} 
 \mathcal{L}_{j}(\theta_j) = ({2n})^{-1}\sum_{i=1}^n\int_{t_j}^{t_{j+1}}
 \left(  
 Y_i^j(t)\left[h_{\theta_j}(Z_i(t),t)-\Bar{h}_{\theta_j}(t)\right]^2dt  -2 \left[h_{\theta_j}(Z_i(t),t)-\Bar{h}_{\theta_j}(t)\right]dN_i^j(t)  
  \right)
\end{align}
where  $$N_i^j(t) = \mathbbm{1} ( X_i\leq t, \delta_i=1, t_j\leq t<t_{j+1}),$$ $$ Y_i^j(t) = \mathbbm{1}  ( X_i \geq t, t_j \leq t< t_{j+1})$$ and we consider $t_0=0$ and $t_{M+1}=\tau$.
If we look more closely, we can see how the counting process $N_i^j(t)$, specific to the intervals $[t_j,t_{j+1})$, is constant outside $[t_j,t_{j+1})$.   Hence, its increment, $dN_i^j(t)$, is null for every subjects $i$ that experiences an event outside that specific interval of time. Moreover, $$
  Y_i^j(t) = \mathbbm{1}  ( X_i \geq t, t_j  \leq X_i< t_{j+1}, t_j \leq t< t_{j+1}) + \mathbbm{1}( X_i \geq t_{j+1}, t_j \leq t< t_{j+1} ).
 $$
Therefore, $Y_i^j(t)$ is a function consistently equal to one that becomes null when the subject experiences the event or is censored.
 Hence, any observation with $X_i<t_j$ doesn't play any role in the j-integral, since $Y^j_i(t)=0$ and $dN_i^j(t)=0$.
However, if $X_i\geq t_{j+1}$, since $dN_i^j(t)=0$ and $Y^j_i(t)=1$, every such observation still appears in the risk set. 
Indeed,  observations that experience the event or are censored after $t_{j+1}$ are still alive in the interval $[t_j,t_{j+1})$ and, therefore, still at risk.

In conclusion, while considering $[t_j,t_{j+1})$interval,  we can censor at $t_{j+1}$ anyone that dies after $t_{j+1}$  and we can eliminate anyone that dies or is censored before $t_{j}$.
More technically, we create therefore for each interval, $[t_0,t_1),\ldots,[t_{M-1},t_M),[t_M,\infty)$, M+1 working ``datasets", 
$$D^j=(X^j_i,\delta^j_i,\tilde{Z}^j_i)_{i=1}^{n_j}$$ 
for $j=0,\ldots,M$, according  to the following principles:
\begin{align}\label{eq:dj}
 X^j_i &=\begin{cases}X_i & t_j\leq X_i<t_{j+1}\\
t_{j+1} & X_i\geq t_{j+1}
\end{cases}, \nonumber
 \\
 \delta^j_i &=\begin{cases}\delta_i & t_j\leq X_i<t_{j+1}\\
0 & X_i\geq t_{j+1}
\end{cases}.
\end{align}
\begin{align}\label{eq:tildeZ}
\tilde{Z}_i^0&=Z_i(t_0) \nonumber \\
\tilde{Z}_i^j&=\Bigl(Z_i(t_j)^\top,\hat{h}_0(\tilde{Z}_i^0),\ldots,\hat{h}_{j-1}(\tilde{Z}_i^{j-1})\Bigl) ^\top .
\end{align}
Here, $n_j =|D_j|$, denotes the cardinality of the at-risk observations, i.e., the set $D_j$.  Note that the at-risk datasets, are rarely of the same size and that typically, $n_0 \geq n_1 \geq \dots \geq n_M$.

The idea  described  above  is inspired by the time-dependent coefficient   survival  models, utilized widely since the early work on  histogram sieves   \citep{murphy1991time} or more generally time-varying coefficient models \citep{hastie1993varying}.   The justification can be understood from breaking down the integrated score  into a  product of time intervals specific score.

\subsection{Estimation} \label{lab:estimation}
Each dataset $D_j$ is now used to estimate the part of the risk $h(Z(t),t)$ that is specific to the interval $j$, that is $h_j(Z(t))$.
To this goal, $M+1$ neural networks, one for each time interval, are constructed. To accommodate the sequential nature of the time, observations within $D_j$ together with outcomes of the trained neural networks from previous time intervals, $\hat h_k (\tilde{Z}^k)$ for $k < j$, are fed into the neural network $j$. The neural network $j$ uses, as loss function, the $j$-th loss $\mathcal{L}_j (\theta_j)$, \eqref{loss2},
relative to that specific interval.
Due to  the assumed structure \eqref{hstruct}, that loss simplifies to:
\begin{align}\label{loss}
& \frac{1}{2n}\sum_{i=1}^{n_j}\int_{t_j}^{t_{j+1}}Y^j_i(t)\left[h_{j,\theta_j}(\tilde Z^j_i)-\Bar{h}_{j,\theta_j}(t)\right]^2dt 
 \nonumber \\
 & \qquad \qquad -
 \frac{1}{n}\sum_{i=1}^{n_j}\int_{t_j}^{t_{j+1}}\left[h_{j,\theta_j}(\tilde Z^j_i)-\Bar{h}_{j,\theta_j}(t)\right]dN^j_i(t) +\lambda \norma{\theta_j}_p
\end{align}
where
\begin{equation}
\Bar{h}_{j,\theta_j}(t)=\frac{\sum_{i=1}^{n_j}h_{j,\theta_j}(\tilde Z^j_i)Y^j_i(t)}{\sum_{i=1}^{n_j}Y^j_i(t)}.
\end{equation}
The above function $\Bar{h}_{j,\theta_j}(t)$ represents the mean of  $h_{j,\theta_j}$ restricted to the risk set at time $t$ which comprises all the subjects still alive.  
Now, noticing that the function $\Bar{h}_{j,\theta_j}(t)$ is a stepwise function that is constant on any interval $[X^j_{r-1},X^j_{r}]$, as shown in the appendix,   the above simplifies to:
\begin{align*} 
& \frac{1}{2n}\sum_{i=1}^{n_j}\sum_{r=1}^{i}\left[h_{j,\theta_j}(\tilde{Z}^j_i)-\Bar{h}_{j,\theta_j}(X^j_r)\right]^2\left(X^j_r-X^j_{r-1}\right)
\\
& \qquad \qquad -
 \frac{1}{n}\sum_{i=1}^{n_j}\left[h_{j,\theta_j}(\tilde{Z}^j_i)-\Bar{h}_{j,\theta_j}(X^j_i)\right]\delta^j_i +\lambda \norma{\theta_j}_p,
\end{align*} 
where $X^j_{0}=t_j$.  See Appendix \ref{a2} for more details.

Here, we notice how the loss cannot be written as a sum of independent individual $i$-specific losses. Indeed, the term $\Bar{h}_{j,\theta_j}(X^j_i)$, as explained before, uses all the individuals still at risk at time $X^j_i$. Thus, the optimization method that relies on breaking down the sample in batches cannot be performed here. This is a common characteristic of every loss related to any continuous survival model. It is the same, for example, in \citet{katzman2018deepsurv}, where the loss used is the partial likelihood that characterizes the Cox proportional model. The application of batch optimization for survival data requires the use in the loss of an approximate risk set, instead of the true one, as explained in \citet{kvamme2019time} where the idea is applied to the Cox model.

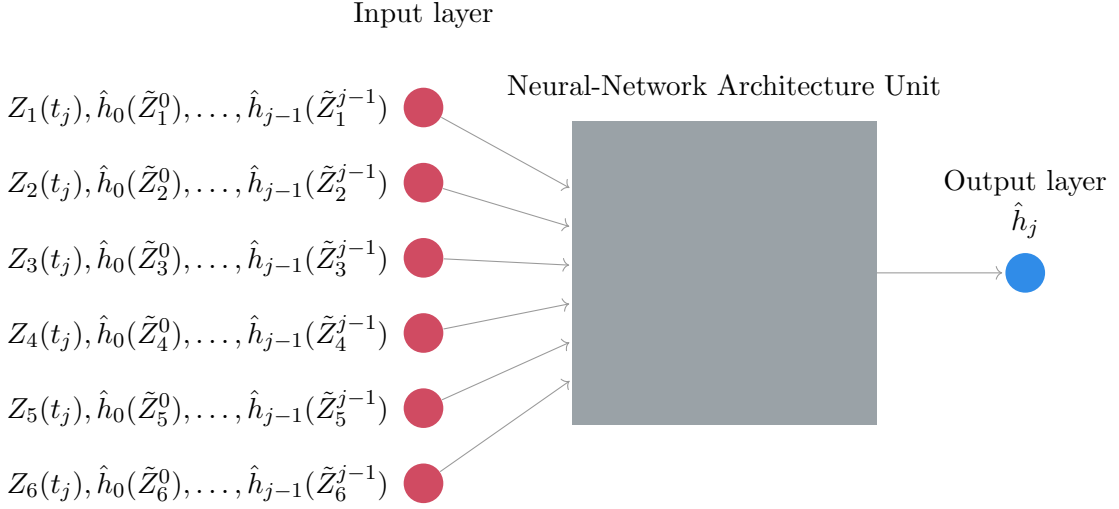
\begin{figure}[h]
\centering
\begin{tikzpicture}[shorten >=1pt,->,draw=black!40, node distance=\layersep]
    \tikzstyle{every pin edge}=[<-,shorten <=1pt]
    \tikzstyle{neuron}=[circle,fill=black!25,minimum size=15pt,inner sep=0pt]
     \tikzstyle{neuron1}=[rectangle,fill=black!15,minimum size=115pt,inner sep=0pt]
    \tikzstyle{input neuron}=[neuron, fill=cardinal!80];
    \tikzstyle{output neuron}=[neuron, fill=bleudefrance];
    \tikzstyle{hidden neuron}=[neuron1, fill=charcoal!50];
    \tikzstyle{annot} = [text width=16em, text centered];

    \foreach \name / \y in {1,...,6} 
        \node[input neuron] (I-\name) at (0,-\y) {};

    
        \path[yshift=1.0cm]
            node[hidden neuron] (H) at (4cm, -4.2cm) {};
    
    \node[output neuron] (O)  at (8cm, -3.2cm){};
  
    \foreach \source in {1,...,6}
            \path (I-\source) edge (H);
            
            
        \path (H) edge (O);          

      \node[annot,left of=I-1, node distance=3cm]  {$Z_1(t_j),\hat{h}_0(\tilde{Z}^0_1),\ldots,\hat{h}_{j-1}(\tilde{Z}^{j-1}_1)$};
       \node[annot,left of=I-2, node distance=3cm]  {$Z_2(t_j),\hat{h}_0(\tilde{Z}^0_2),\ldots,\hat{h}_{j-1}(\tilde{Z}^{j-1}_2)$};
        \node[annot,left of=I-3, node distance=3cm]  {$Z_3(t_j),\hat{h}_0(\tilde{Z}^0_3),\ldots,\hat{h}_{j-1}(\tilde{Z}^{j-1}_3)$};
         \node[annot,left of=I-4, node distance=3cm]  {$Z_4(t_j),\hat{h}_0(\tilde{Z}^0_4),\ldots,\hat{h}_{j-1}(\tilde{Z}^{j-1}_4)$};
          \node[annot,left of=I-5, node distance=3cm]  {$Z_5(t_j),\hat{h}_0(\tilde{Z}^0_5),\ldots,\hat{h}_{j-1}(\tilde{Z}^{j-1}_5)$};
           \node[annot,left of=I-6, node distance=3cm]  {$Z_6(t_j),\hat{h}_0(\tilde{Z}^0_6),\ldots,\hat{h}_{j-1}(\tilde{Z}^{j-1}_6)$};

       \node[annot,above of=H, node distance=2.5cm] (hl) {Neural-Network Architecture Unit};
         \node[annot,above of=O, node distance=0.7cm]  {$\hat{h}_j$} ;
        \node[annot,above of=O, node distance=1.2cm] (O) {Output layer};
        \node[annot,above of=I-1, node distance=1.0 cm] {Input layer \phantom{gkfjnkjfngkdjgkdjgnkjfhkgjdhfdfjbgdfkjnfg}};
         \node[annot,right of=hl1] {};  
         
\end{tikzpicture}
\caption{Deep Hazard: An example of $j$-th NN  with Input Layer consisting of six still at risk observations and forward passes of previously learned $j-1$ networks,  with arbitrary NN Architecture  Unit.  }\label{fig:3}
\end{figure}

 In our experiments, every hidden fully-connected layer is followed by a nonlinear activation function and a dropout layer; however, the method can take any architecture of the layers of interest.  As is common in neural networks, the output is a plain weighted combination of the last hidden layer's output. No activation function is used for the computation of the output. An example of the input structure of each $j$-th NN network is depicted in Figure \ref{fig:3}. We observe how each set of still-at-risk observations is enriched with additional features coming out of feed-forward passes run on previously fitted $j-1$ networks.
 
\begin{figure}[h]
\centering
\begin{tikzpicture}[shorten >=1pt,->,draw=black!40, node distance=1.2cm]
    \tikzstyle{every pin edge}=[<-,shorten <=1pt]
    \tikzstyle{neuron}=[circle,fill=black!25,minimum size=18pt,inner sep=0pt]
    \tikzstyle{input neuron}=[neuron, fill=cardinal!80];
    \tikzstyle{output neuron}=[neuron, fill=bleudefrance];
    \tikzstyle{hidden neuron}=[neuron, fill=white];
    \tikzstyle{annot} = [text width=8em, text centered]
     \tikzstyle{layerbox}=[square,fill=black!25,minimum size=25pt,inner sep=0pt]
  
    \node[output neuron,pin={[pin edge={<-}]left:Input}] (O) {};
    
    \node[output neuron] (O1) at (2.7cm,0cm) {};
    \node[output neuron] (O2) at (5.4cm, 0cm){};
   
      \node[output neuron,pin={[pin edge={->}]right:Output}] (O3) at (10cm, 0cm){};
        \node[hidden neuron] (O4) at (6.7cm, 0.5cm){};
        
 \draw [annot] (O) edge node[above] {\tiny $a(w_1h_1+b_1)$} (O1) ;
\draw (O) to [bend left] (O2);
\draw   (O) to [bend left] (O3);
 \draw [annot] (O1) edge node[above] {\tiny $a(w_2h_2+b_2)$} (O2) ;
\draw  (O1) to  [bend left] (O3);
 \draw [annot] (O2) edge node[above] {\tiny $a(w_{\tiny{3}}h_{\tiny{3}}+b_{\tiny{3}})$} (O3) ;
  \node[annot,below of=O, node distance=0.7cm]  {$h_0$} ;
 \node[annot,above of=O1, node distance=0.7cm]  {\tiny $a(w_2h_1+b_2)$} ;
  \node[annot,below of=O1, node distance=0.7cm]  {$h_1$} ;
   \node[annot,above of=O2, node distance=1.4cm]  {\tiny $a(w_3h_1+b_3)$} ;
 \node[annot,below of=O2, node distance=0.7cm]  {$h_2$} ;
    \node[annot,above of=O4, node distance=0.5cm]  {\tiny $a(w_3h_2+b_3)$} ;
  \node[annot,below of=O3, node distance=0.7cm]  {$h_3$} ;

     \node[input neuron] (D) at (0cm,-1.6cm) {};
       \node[annot,below of=D, node distance=0.7cm]  {$D_0$} ;
      \node[input neuron] (D2) at (2.7cm,-1.6cm) {};
          \node[annot,below of=D2, node distance=0.7cm]  {$D_1$} ;
           \node[input neuron] (D3) at (5.4cm,-1.6cm) {};
               \node[annot,below of=D3, node distance=0.7cm]  {$D_2$} ;
                \node[input neuron] (D4) at (10cm,-1.6cm) {};
                    \node[annot,below of=D4, node distance=0.7cm]  {$D_3$} ;
  
  \draw  (D) to(O);
  \draw [britishracinggreen!80,dotted] (D) to  (D2);
    \draw  (D2) to(O1);
  \draw [britishracinggreen!80,dotted] (D2) to  (D3);
      \draw  (D3) to(O2);
  \draw [britishracinggreen!80,dotted](D3) to  (D4);
     \draw  (D4) to(O3);
\end{tikzpicture}
\caption{Deep Hazard:  unpacked time-convolutions. Each time-specific neural network (NN)  disregards hidden layers. The blue nodes denote outputs of time-specific NNs whereas, arrows denote feed-forward interactions over time. At each arrow, we show the activation function $a$ and weights $w_i$ and biases $b_i$, $i=1,2,3.$. Dotted lines denote dependence of still-at-risk individuals comprising inputs of each time-specific NN.
 }\label{fig:2}
\end{figure}
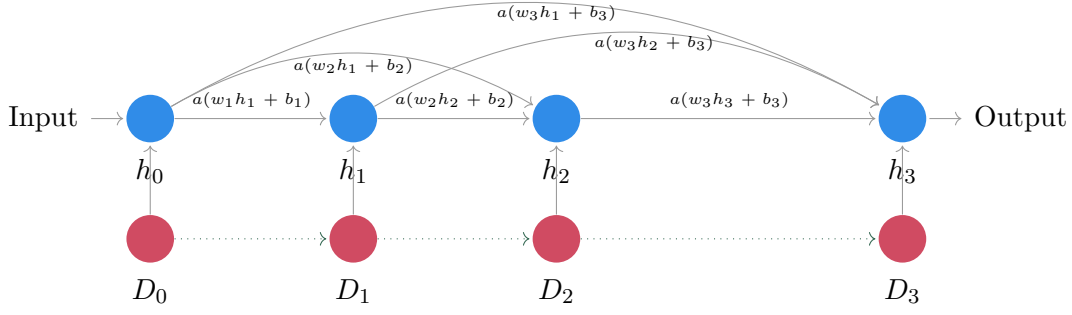

A possible time-convolution, as proposed earlier, can be visualized in Figure \ref{fig:2}. 
There, we illustrate how the time-dependent outputs of each $j$-th NN are passed onto all of the future NNs. Moreover, we indicate that each dataset, $D_j$, comprising inputs of $j$-th NN, depends on the previous dataset $D_{j-1}$.

After running $M+1$ sequential (or time-convoluted) neural networks on the $M+1$ working datasets, we obtain the optimized weights and biases, denoted here with  $\theta_0,\dots,\theta_M$. 
To form prediction of the new, test individual  we proceed as follows.  With a little abuse in notation, let 
$$Z(t)=\left\{Z(t_0),\ldots,Z(t_M)\right\}$$
 be observations pertaining to a sequence of follow up visits, at times $t_0,\dots,t_M$, of a new patient. For each $\theta_0,\cdots, \theta_M$, 
with  standard forward pass (evaluating the estimated hazard for a specific new observation), one for each network, we get the following estimates:
$$\hat{h}_0(\tilde Z^0):=h_{\theta_0}(\tilde{Z}^0) , \ldots, \hat{h}_M(\tilde Z^M):=h_{\theta_M}(\tilde{Z}^M) .$$
Here, each $\tilde Z$ is constructed following equation \eqref{eq:tildeZ}, in other words, each prediction of the risk  at a future time $t_j$ uses the formed predictions of previous time points $t_0,\cdots, t_{j-1}$.
Combining these into a single risk estimator is then simple.  Following \eqref{model} we obtain for an out-of-sample individual
\begin{equation}\label{eqh}
\hat{h}(Z(t),t) = \hat{h}_0(\tilde Z^0) \mathbbm{1}(t\leq t_1) + \hat{h}_1(\tilde{Z}^1) \mathbbm{1} (t_1<t\leq t_2) + \cdots + \hat{h}_{M}(\tilde Z^M)\mathbbm{1}(t> t_M)
\end{equation}

For more details on the training process see Algorithm \ref{alg:1}. 

 \begin{algorithm}[H]\caption{DeepHazard:Training}\label{alg:1}
\begin{algorithmic} 
\Require Training set 
$
(X_i,\delta_i,Z_i(t_0),\ldots,Z_i(t_M))_{i=1}^n
$, 
hyper parameters  and hidden layers   of $M+1$ neural networks

 \State $\theta$ $\leftarrow$ initialize   weights and biases  
\State Set $\tilde Z_i^0 \leftarrow Z_i (t_0) $
 \State Create $D_0$ dataset according to \eqref{eq:dj}
 \State $\theta_0$ $\leftarrow$  neural network  initialized at $\theta$ and with input $\tilde Z_1^0, \dots, \tilde Z_n^0$

\For{$j$  in $0:M$ }
\State $\theta$ $\leftarrow$ initialization of  weights and biases  accordingly to initialization method
  \State Set $\tilde Z_i^j\leftarrow \Bigl(Z_i(t_j),\hat{h}_0(\tilde{Z}_i^0),\ldots,\hat{h}_{j-1}(\tilde{Z}_i^{j-1})\Bigl)$
 \State Create $D_j$ dataset according to \eqref{eq:dj}
 \State Set $n_j = \mbox{card}(D_j)$
 \State $\theta_j$ $\leftarrow$  neural network  initialized at $\theta$ and with input $\tilde Z_1^j, \dots, \tilde Z_{n_j}^j$
 \For {$i$ in $1 : n$}
\State $ \hat{h}_j(\tilde{Z}^j_i) \leftarrow h_{\theta_j}(\tilde{Z}_i^j) $   \Comment{forward pass of the $j$-th NN on the training data}
\State $r_i\leftarrow\sum_{l=1}^nY_l(X_i)$ \Comment{number of people at risk at that time}
\State $\bar{h}_j(X_i)\leftarrow {\sum_{l=i}^n r_i^{-1} \hat{h}_j(\tilde{Z}^j_l) }$ 
\EndFor
\EndFor 
 
\For{$i$ in $1:n$}
\State   $J_i\leftarrow \left\{j\;:\:t_{j}\leq X_i<t_{j+1}\right\}$ \Comment{which interval contains the censored time}
\State   $r_i\leftarrow\sum_{l=1}^nY_l(X_i)$
\State $$\hat{\Lambda}_0(X_i)\leftarrow \sum_{l=1}^i\frac{\delta_l}{r_l}-\sum_{j=0}^{J_i}\sum_{\ s : t_j\leq X_s <t_{j+1}}[X_{s+1}-X_{s}]\bar{h}_j(X_s)
$$
\EndFor 

\Return   

A matrix 
$
\hat{h} =[\hat{h}_j(\tilde{Z}^j_i)]_{i=1,\ldots,n; j=1,\ldots,M}
$

The vectors $ \theta_0,\ldots,\theta_M $ \Comment{weights and biases for each neural network}

A vector $\Bigl(\hat{\Lambda}_0(X_1),\ldots,\hat{\Lambda}_0(X_n)\Bigl) $
 
\end{algorithmic}
\end{algorithm}

\subsection{Prediction of the survival} \label{lab:prediction}

Practitioners are often concerned with predicting the survival rate of a new patient for a given period of time in the future: survival at one, five, twenty years after diagnosis, for example. With estimated risks of the previous section, we only need to design baseline estimates of the hazard.  When considering   an additive hazards model $\lambda(t| D,Z)=\lambda_0(t)+\beta Z(t)$ as explained in \citet{lin1994semiparametric}, a semiparametric estimate of the cumulative baseline can be proposed. Here, we directly extend their semi-parametric approach.

 Observe that under the model \eqref{model},  
\[
dN_i(t) = d M_i(t) + \int_{0}^t  Y_i(u) d \Lambda_0 (u )   +  \int_{0}^t Y_i(u)  h (Z_i(u),u) du
\]
it is natural to consider  the following estimator
\begin{equation}\label{lambda0}
\hat{\Lambda}_0(t)=\int_0^t\frac{\sum_{i=1}^n\left\{dN_i(u)-Y_i(u)\hat{h}( Z_i(u),u)du\right\}}{\sum_{i=1}^nY_i(u)},
\end{equation}
with $\hat h$ as defined in \eqref{eqh}. Our time-convolutions provide a way to also estimate cumulative baseline hazards for each time-interval. Therefore our method allows data exploration in each time interval as well as overall. We show the equivalence of the two approaches in the Appendix \ref{sec:3}.





 Lastly, we compute  the predicted survival curve by combining the results of the $M$ neural network predictions in the following way
\begin{equation}\label{eqS}\hat{S}(t\mid Z(t))=\begin{cases}
\exp(-\hat{\Lambda}_0(t)-\hat{h}_0(\tilde{Z}^0)t) & t<t_1\\
\exp(-\hat{\Lambda}_0(t)-\hat{h}_1(\tilde Z^1)(t-t_1)-\hat{h}_0(\tilde Z^0)t_1) & t_1\leq t<t_2\\
\exp(-\hat{\Lambda}_0(t)-\hat{h}_2(\tilde Z^2)(t-t_2)-\hat{h}_1(\tilde Z^1)(t_2-t_1)-\hat{h}_0(\tilde Z^0)t_1) & t_1\leq t<t_2\\
\ldots
\end{cases}.
\end{equation}
 Here, we take as value for $Z(t)=Z(t_{J_t})$ where $J_t=\{j\;:\;t_{j}\leq t <t_{j+1}\}$.
 
Finally we construct the following adjusted version of the predicted survival
$$
\hat{S}(t\mid Z(t))=\min_{s\leq t}\hat{S}(s\mid Z(s))
$$ 
guaranteeing the estimator of the survival to be decreasing, consequently  avoiding the well known problem of possibly negative risk and therefore hazard. For more details see Algorithm \ref{alg:2}.


  

\begin{algorithm}[h]
\caption{DeepHazard:Prediction}\label{alg:2}
\begin{algorithmic}
\Require Test data $\mathbb{Z}(t_0),\ldots,\mathbb{Z}(t_M)$, event times from the Test data, $\{ X_1,\dots, X_n \}$  and outcomes of DeepHazard:Training, i.e., $\theta_0,\ldots,\theta_M$ and $\hat{\Lambda}_0(X_1),\ldots,\hat{\Lambda}_0(X_n)$

\State Set $\tilde{\mathbb{Z}}^0\leftarrow \mathbb{Z}(t_0)$ 
\For{$j$ in $0:M-1$}
\State $\hat{h}_j(\tilde{\mathbb{Z}}^j) \leftarrow h_{\theta_j}(\tilde{\mathbb{Z}}^j)$  \Comment{forward pass of the DeepHazard on the test data}
\State $\tilde{\mathbb{Z}}^{j+1}\leftarrow \left(\mathbb{Z}(t_{j+1}),\hat{h}_0(\tilde{\mathbb{Z}}^0),\ldots,\hat{h}_{j}(\tilde{\mathbb{Z}}^{j})\right)$ 
 \EndFor  
 \State  $\hat{h}_M(\tilde{\mathbb{Z}}^M) \leftarrow h_{\theta^M}(\tilde{\mathbb{Z}}^M)$   
\For{$i$ in $1:n$}
\State Set $J_i \leftarrow \left\{j\;:\;t_j\leq X_i <t_{j+1}\right\}$  \Comment{Here $\mathbb{Z}(X_i)\approx\mathbb{Z}(t_{J_i})$ }
\State Set 
$$\hat{S}(X_i \mid \mathbb{Z}(X_i)) \leftarrow \exp \left (-\hat{\Lambda}_0(X_i)-\hat{h}_{J_i}(\tilde {\mathbb{Z}}^{J_i})(X_i-t_{J_i})-\sum_{l=0}^{J_i-1}\hat{h}_{l}(\tilde{\mathbb{Z}}^l)(t_{l+1}-t_l) \right)$$
\State Set $\hat{S}(X_i \mid \mathbb{Z}(X_i)) \leftarrow  \min_{l\leq i}\hat{S}(X_l \mid \mathbb{Z}(X_l))$ \Comment{monotonicity guarantee}
\EndFor

\Return  $\hat{S}(X_1\mid \mathbb{Z}(X_1)),\ldots,\hat{S}(X_n\mid \mathbb{Z}(X_n))$
\end{algorithmic}
\end{algorithm}

Until now, we have implicitly assumed that, any new observation will have $Z(t)$ measured at the same time points used to train the network - $t_0,\ldots,t_M$. If this is not the case we approximate $Z(t_j)$ with the nearest $Z(t)$ available. More in details, if the measurements $Z(\tilde{t}_0),\ldots,Z(\tilde{t}_{\tilde M})$ are available, we make use of the following approximation: $Z(t_i)=Z(\tilde{t}_{J_i})$ where $J_i=\argmin_{j=1,\ldots,\tilde{M}}|t_i-\tilde{t}_j|$.

We study in simulation the effect of the number and the placement of the time points that define the $M+1$ intervals. We show that the performance of our procedure remains stable when the time points at which the covariates are measured shift or more time points are added. The only restriction that needs to be kept in mind is that we need to have enough observations to train the last neural network, that, we remind, uses as input only the observation still at risk after $t_M$. The last time point therefore cannot be too large in comparison to the magnitude of the censored event time of our sample.

\section{Finite sample  experiments}\label{lab:3}
 
In this section, we evaluate the performance of Deep Hazard in finite samples.
We compare DeepHazard with the Additive Hazards Model, \citep{aalen1980model}, that  presuposes $$\lambda(t \mid Z(t))=\lambda_0(t)+\beta(t)Z(t),$$  and with the Time dependent Cox Model, \citep{fisher1999time}, that assumes $$\lambda(t \mid Z(t))=\lambda_0(t)\exp{(\beta Z(t))}.$$  
We use the R packages \emph{Timereg} and \emph{Survival}, respectively to fit the Additive Hazards Model and the Time dependent Cox model.
As a measure of performance, we use the time dependent C-index as proposed by \cite{antolini2005time},
\begin{align*}
\mbox{C}_{\mbox{\tiny index}}&= \sum_{i=1}^n\sum_{j=1;j\neq i}^n\mathbbm{1}\Bigl(\hat{S}(t_i \mid Z_i(t))<\hat{S}(t_j \mid Z_j(t))\Bigl)
p_{i,j},
\end{align*}
where
\begin{align*}p_{i,j}&=\frac{ \mathbbm{1}\left\{t_i <t_j,\;\delta_i=1\right\}+\mathbbm{1}\left\{t_i =t_j,\;\delta_i=1,\;\;\delta_j=0\right\}}{\sum_{i=1}^n\sum_{j=1;j\neq i}^n\left[\mathbbm{1}\left\{t_i <t_j,\;\delta_i=1\right\}+\mathbbm{1}\left\{t_i =t_j,\;\delta_i=1,\;\;\delta_j=0\right\}\right]}.
\end{align*}

We also introduce a new measure, the integrated mean square prediction error (IMSPE), defined as follows:
\begin{equation*}
\mbox{IMSPE}=\frac{1}{\tau}\int_0^\tau \frac{1}{n}\sum_{i=1}^n\left\{\hat{S}(t \mid Z(t))-S(t \mid Z(t))\right\}^2
\end{equation*}
to capture the quality of the  prediction error through time.

We implement our neural network in PyTorch, \url{https://github.com/deniserava/DeepHazard}.
The implementation is flexible in that the user can choose the structure of the Neural Network: the number of hidden layers, number of hidden nodes, activation function, and a dropout rate. Moreover, the following Hyperparameters related to the optimization procedure of the neural networks, as initialization method, optimizer used, learning Rate (lr), number of Epochs (E), and early stopping can be chosen. The user can also select the regularization parameters $\lambda$ and $p$ of the loss \eqref{loss}.
It is worth noting that Epochs are updating the network weights and biases (parameters) through a suitable optimization method, but stay un-permuted to preserve the order of the survival outcomes. 
A list of the popular activation functions, that we implemented, can be found in Appendix \ref{a1}; see Table \ref{act}.

Our numerical experiments are evaluated on simulated data.
We focus on the settings with time-varying covariates. There is a need to describe data-generating processes for the hazard models in the presence of time-varying covariates.
 The latter are generated using the procedure described in Algorithm \ref{alg:3}.

\begin{algorithm}[h]
\caption{Time-dependent Simulation}\label{alg:3}
\begin{algorithmic}
\Require  Covariate function $z$ such that $Z(t) = z(Z,t)$ and $Z$ follows distribution  $\mathcal{Z}$, hazard function $h(\cdot, \cdot)$, baseline hazard $\lambda_0(\cdot)$, censoring level $\ell$, follow up times $t_1,\dots, t_M$, sample size $n$
\For {$i=1,\dots, n$}
\State Simulate $\omega$ from Uniform distribution $U(0,1)$
\State Let $Z$ be a realization of a random draw from $\mathcal{Z}$.
\State  Let $T_i = t$ where $t$ solves  
\[
f(t)=\omega
\] 
\indent where  $Z(u) := z(Z,u)$
\[
f(t) :=  \exp \left[ -  \int_0^t \left\{\lambda_0 (u) + h(u, Z(u))\right\} du \right]
\]
\Comment{$f(t)$ stands to denote the function $S(t|Z(t))$}
 \For {$j=1,\cdots, M$}
\State  Let $Z_i (t_j)=z(Z,t_j)$ 
\EndFor
\EndFor
 \State  Simulate $n$ independent censoring time $C_i$ from Uniform distribution $U(0,c)$  \\
 \Comment{$c$ is such that censoring level is below some level $\ell$}
\State Set $X  = \min \{ T ,C \}$ \\
 \Comment{Observed censored event times}
 \State Let $\delta  =\mathbbm{1}\{T  \leq C \}$
 \Comment{Observed censoring indicator}

\Return  Data $ \left\{X_i, \delta_i, Z_i(t):=(Z_i(t_1),\ldots,Z_i(t_M))\right\}_{i=1}^n$
\end{algorithmic}
\end{algorithm}

\subsection{ Impact of the sample size}

We assume the data is generated according to the following four different hazards models.  Below '$*$' denotes multiplication.
Model 1 follows additive structure but the covariates are highly correlated and non-linear. Model 2 considers further interactions with time whereas Model 3 works with highly non-linear interactions. Model 4 is perhaps the most challenging one. 
\begin{itemize}
\item[]{Model 1: $$\lambda(t\mid Z)=4t^3+Z_1(t)*Z_2(t)+Z_1(t)*Z_3(t)+Z_1(t)*Z_3(t)*Z_2(t)$$}
\item[]{Model 2: $$\lambda(t\mid Z)=4t^3+\cos(t)[Z_1(t)*Z_2(t)]+ | \log(t+1)| Z_1(t)*Z_2(t)+t^3Z_3(t)^2$$}
\item[]{Model 3: \begin{align*}
\lambda(t\mid Z)&=4t^3+\cos(t)[Z_1(t)*Z_2(t)]+\mid \log(t+1)\mid Z_1(t)Z_2(t)+t^3Z_3(t)^2
\\
& \qquad +\cos[Z_1(t)*Z_3(t)]+Z_1(t)*Z_3(t)+\frac{1+t^2}{t+1}Z_1(t)*Z_2(t)+Z_1(t)^3*Z_2(t)^4
\end{align*}
}
\item[]{Model 4: $$\lambda(t\mid Z)=4t^3+\frac{1}{t+1}Z_1(t)*Z_2(t)+\frac{1}{Z_1(t)*Z_2(t)*Z_3(t)^2+1}$$}
\end{itemize}

The covariates are generated according to the following structure
 \begin{equation} \label{eq:Z}
 Z_i(t)=\begin{cases} \sqrt{t} \ Z_{0i} & t\leq 0.6\\ \sqrt{0.6} \ Z_{0i} & \mbox{otherwise} \end{cases}
 \end{equation}  where $i=1,2,3$ and $Z_{0i}\sim U(0,20)$ for $i=1,2,3$ except for {\bf Model 1}, where $Z_{01}\sim U(0,10),Z_{02}\sim U(0,20),Z_{0i}\sim U(0,30)$. 
 
 We assume to measure the covariates at the following times $0.001,0.2,0.4,0.6$. We generate 1000 observations for the training set and for the test set. We fit to the training set the Additive Hazards Model, the Time-dependent Cox Model and DeepHazard. 1000 epochs are used with early stopping rate $1e^{-5}$ and initialization method he Normal is employed. The C-index of each Model is presented in Table \ref{result1}. The Hyperparameters chosen for our neural network are reported in Table \ref{ti1}. 
 
 We report also the  Oracle C-index that uses the true $S(t \mid Z(t))$ for comparison purposes. We then repeat the simulations with a sample size of 200 for both train and test set.  We observe superior performance of DeepHazard both across samples as well as Models. Moreover, C-index is often extremely close to the oracle C-index indicating  certain optimality.
  
\begin{table}[H] \centering \caption{Result of Simulation for additive Hazards Model, Time-dependent Cox and our method (DeepHazard) for Model 1, 2 , 3 and 4. }\label{result1}

\begin{tabular}{@{\extracolsep{5pt}} lcccc} 
\Xhline{.8pt}  
  \\[-3ex] 
 & Model 1 & Model 2 & Model 3 & Model 4\\
 \Xhline{.8pt} 
  \rowcolor{lightgray}\multicolumn{5}{c}{ C-index} \\ \Xhline{.8pt} 
 $n=1000$\\
Oracle & $0.765$ & $0.749$ & $0.716$ & $0.742$\\
\cline{2-2} \cline{3-3} \cline{4-4} \cline{5-5}
Deep Hazard  & \cellcolor{charcoal!60}$\cellcolor{charcoal!60}0.752$& \cellcolor{charcoal!60}$0.735$ & \cellcolor{charcoal!60}$0.716$ &\cellcolor{charcoal!60}$0.733$\\
Additive Hazards    & $0.665$    & $0.590$ & $0.674$   & $0.636$\\
Time-dependent Cox & $0.726$& $0.718$ & $0.703$ & $0.717$\\
\hline \\[-1.8ex] 
$n=200$ \\
Oracle & $0.743$ & $0.734$ & $0.681$ & $0.739$ \\
\cline{2-2} \cline{3-3} \cline{4-4} \cline{5-5}
Deep Hazard  & \cellcolor{charcoal!60}$0.726$& \cellcolor{charcoal!60}$0.717$ & $0.666$ & \cellcolor{charcoal!60}$0.727$\\
Additive Hazards    & $0.635$    & $0.174$ & $0.651$   & $0.598$\\
Time-dependent Cox  & $0.713$& $0.700$ & \cellcolor{charcoal!60}$0.676$ & $ 0.699$\\
\end{tabular} 
\end{table}

\begin{table}[H] \centering \caption{DeepHazard experimental Hyperparameters of Table~\ref{result1}.}
\begin{tabular}{@{\extracolsep{5pt}} lcccc} \label{ti1}{\it Hyperparameter} &  & &  & \\
\hline
$n=1000$ & Model 1 & Model 2 & Model 3 & Model 4\\
Optimizer  &{\it Adam}& {\it Adam} & {\it Adam} & {\it Adam}\\
Activaction   & {\it Elu}$(0.1)$    & {\it Relu}  & {\it Elu}$(0.1)$/{\it Selu}   & {\it Selu} \\
N. Dense Layer   & $5$& $2$ & $2$ & $ 2$\\ 
N. Nodes Layer & 10/15/20/15/10 & 10 & 20 & 10\\
Learning rate & $0.01$ & $2e-2$ & $2e-1$ & $2e-1$ \\
$\lambda$  & $1e-5$ & $1e-3$ & $1e-5$ & $1e-5$\\
Penalty & Ridge & Ridge & Ridge & Ridge \\
Dropout   & $0.2$    & $0.2$ & $0.2$   & $0.2$\\
\hline \\[-1.8ex] 
$n=200$ 
\\
Optimizer  &{\it Adam}& {\it Adam} & {\it Adam} & {\it Adam}\\
Activaction   & {\it Selu}     & {\it Relu}  & {\it Selu}    & {\it Relu} \\
N. Dense Layer   & $2$& $2$ & $3$ & $ 2$\\ 
N. Nodes Layer & 10 & 10 & 10/15/10 & 10\\
Learning rate & $2e-1$ & $2e-2$ & $1e-3$ & $2e-1$ \\
$\lambda$  & $1e-2$ & $0.41$ & $0.61$ & $1e-4$\\
Penalty & Ridge & Ridge & Ridge & Ridge\\
Dropout   & $0.2$    & $0.2$ & $0.1$   & $0.2$\\
\end{tabular} 
\end{table}

\begin{figure}[H]
    \centering
    \begin{subfigure}[]{0.32\textwidth}
        \centering
     \includegraphics[width=0.95\textwidth]{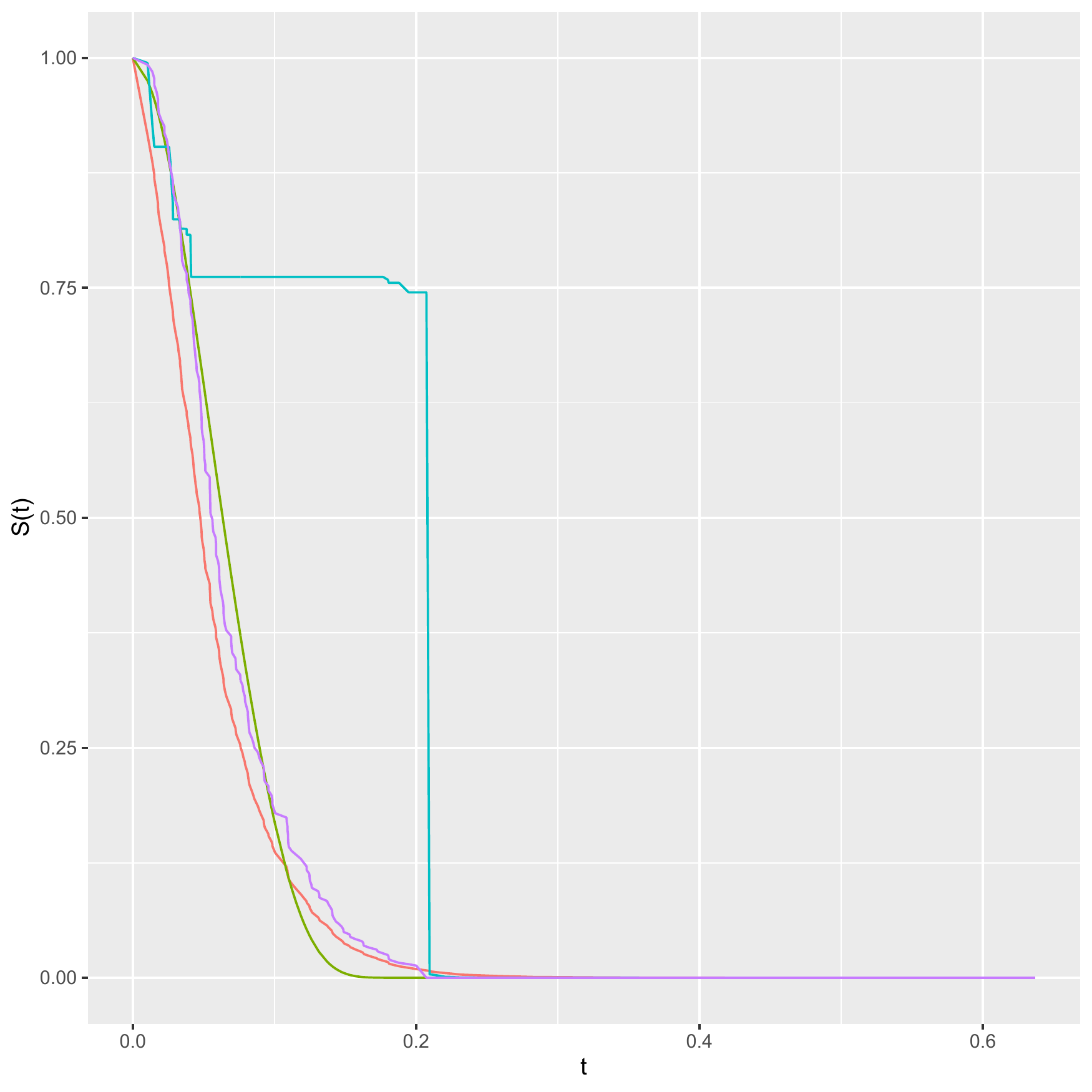}
        \caption{High}
    \end{subfigure}%
    \begin{subfigure}[]{0.32\textwidth}
        \centering
    \includegraphics[width=0.95\textwidth]{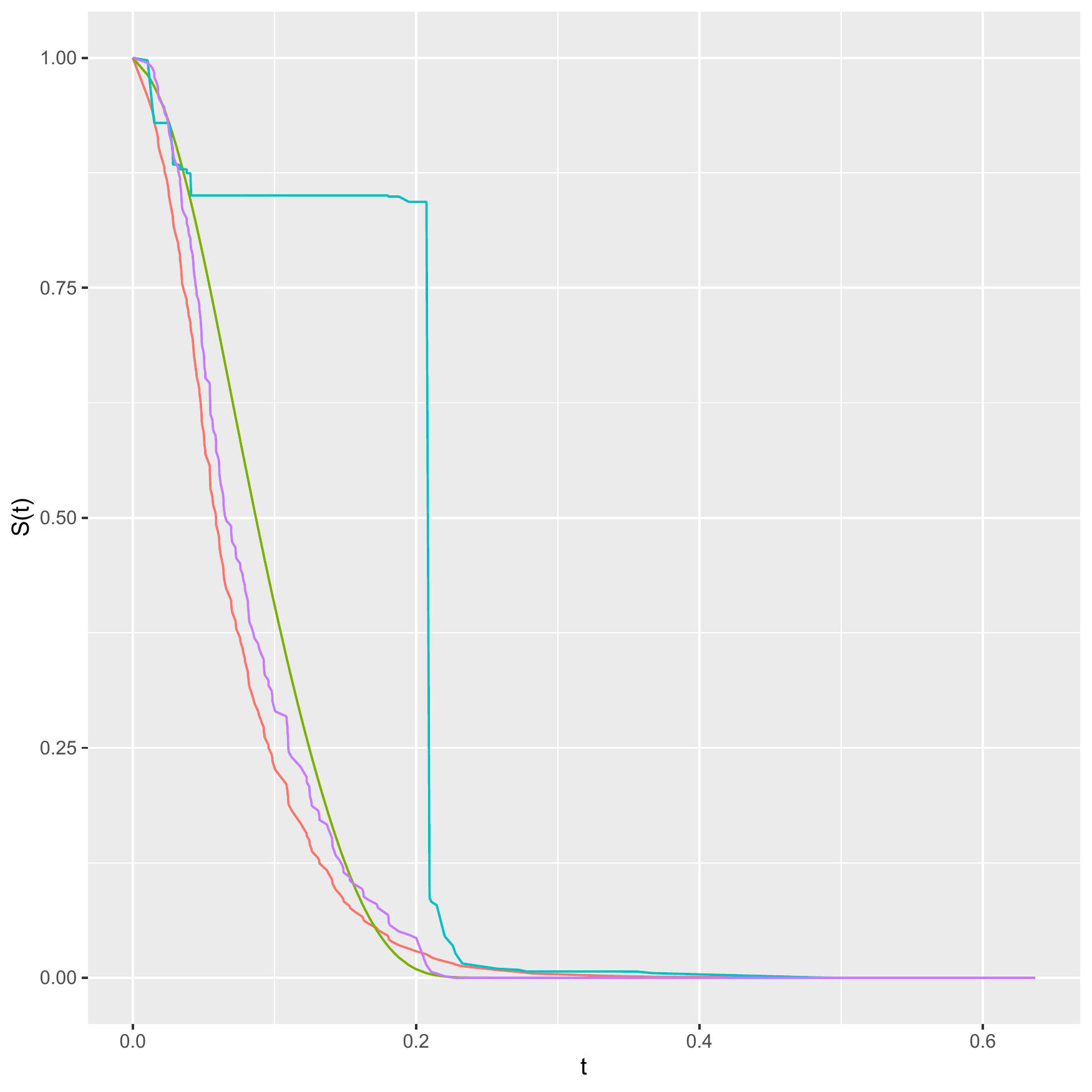}
        \caption{Median-high}
    \end{subfigure}
        \begin{subfigure}[]{0.32\textwidth}
        \centering
     \includegraphics[width=0.95\textwidth]{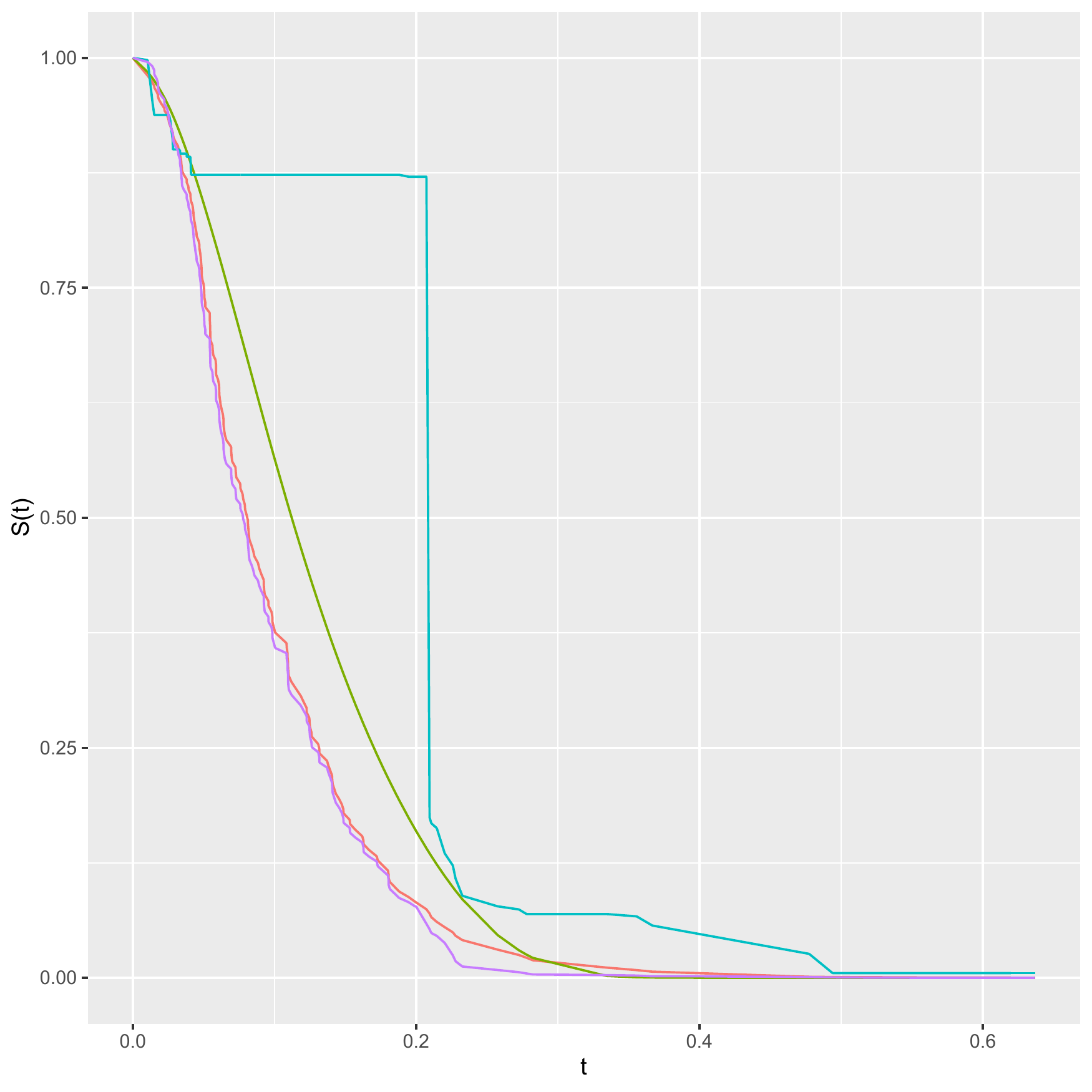}
        \caption{Median-low}
    \end{subfigure}%
    \begin{subfigure}[]{0.32\textwidth}
        \centering
    \includegraphics[width=0.95\textwidth]{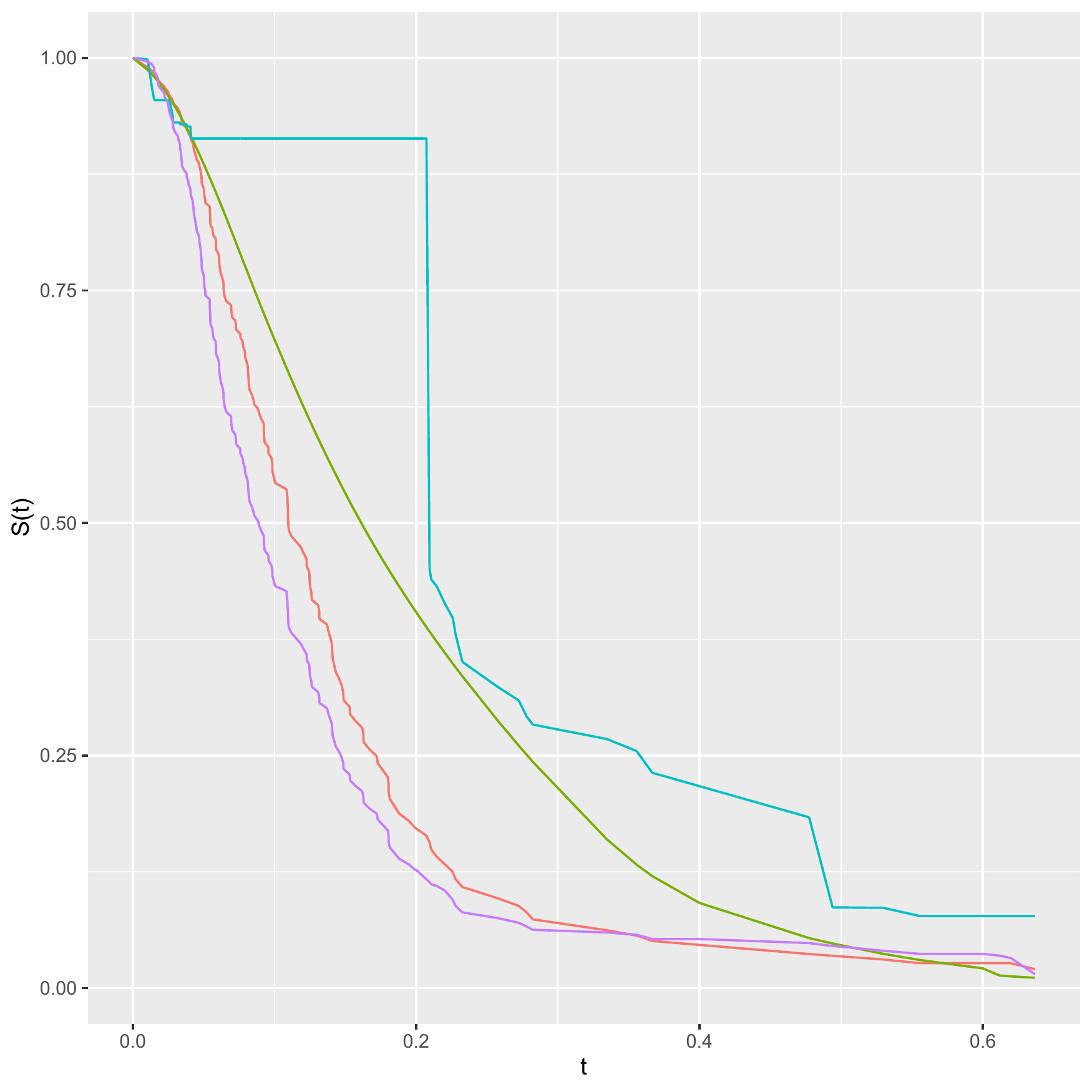}
        \caption{Low}
    \end{subfigure}
    \centering
    \begin{subfigure}[]{0.32\textwidth}
        \centering
     \includegraphics[width=0.95\textwidth]{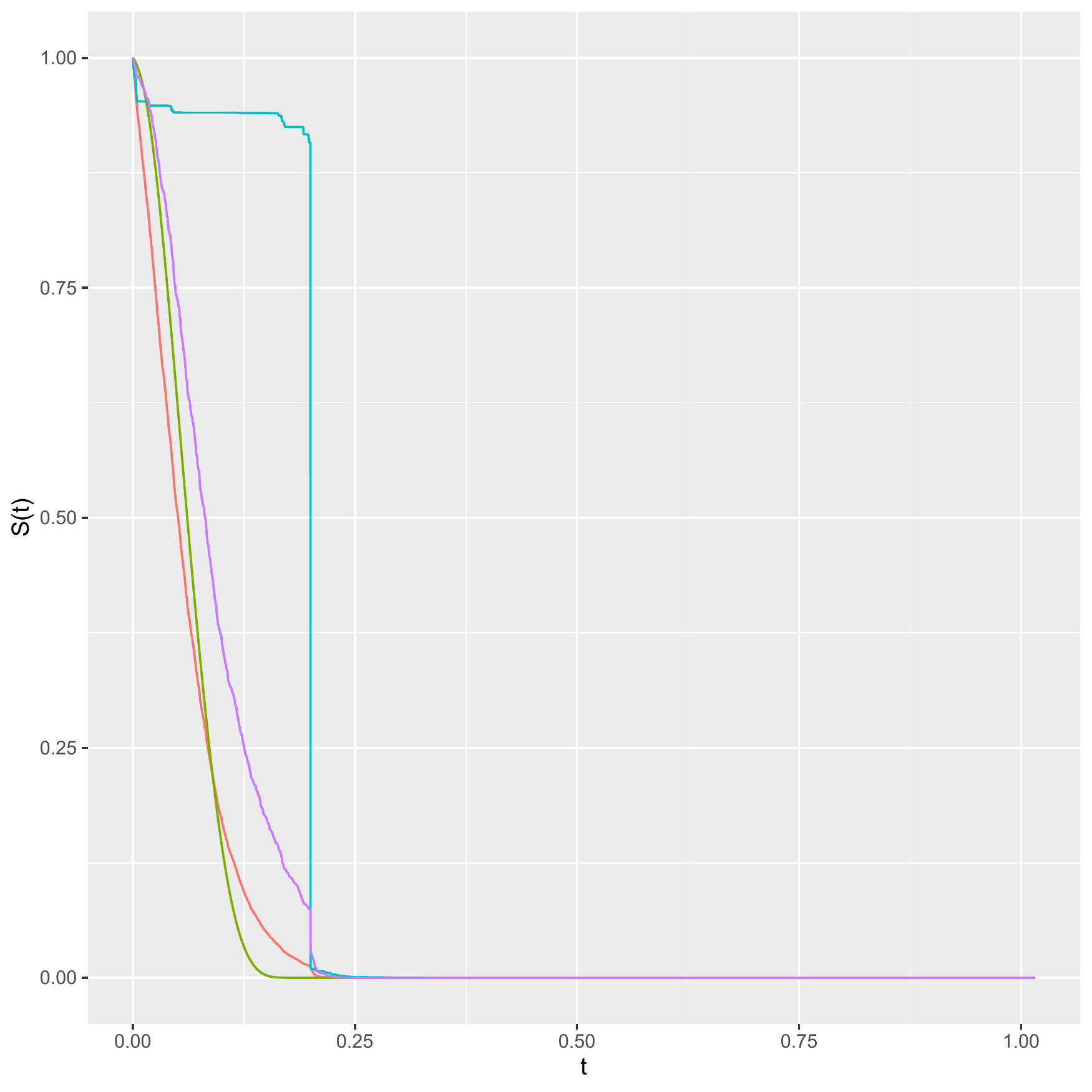}
        \caption{High}
    \end{subfigure}
    \begin{subfigure}[]{0.32\textwidth}
        \centering
    \includegraphics[width=0.95\textwidth]{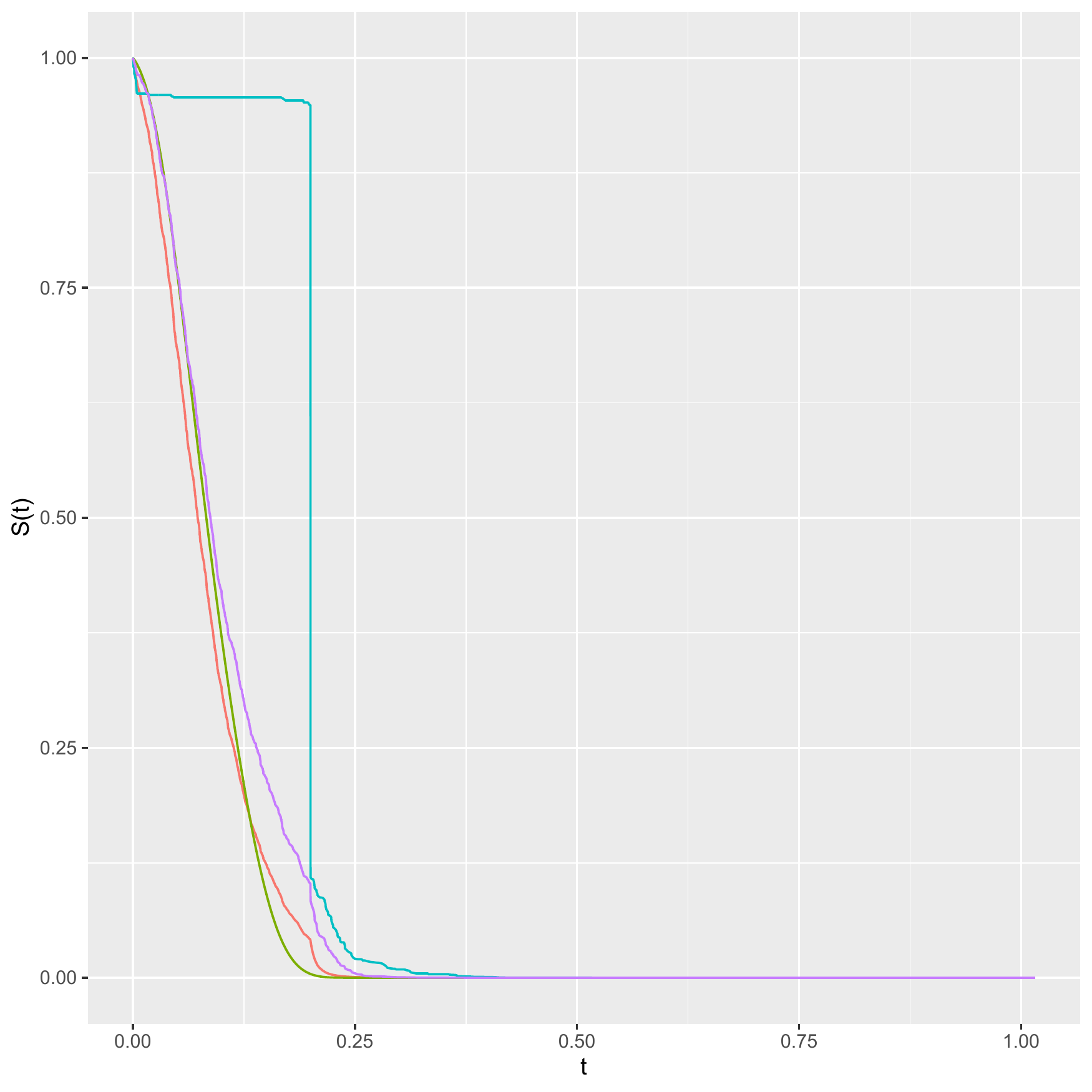}
        \caption{Median-high}
    \end{subfigure}
        \begin{subfigure}[]{0.32\textwidth}
        \centering
     \includegraphics[width=0.95\textwidth]{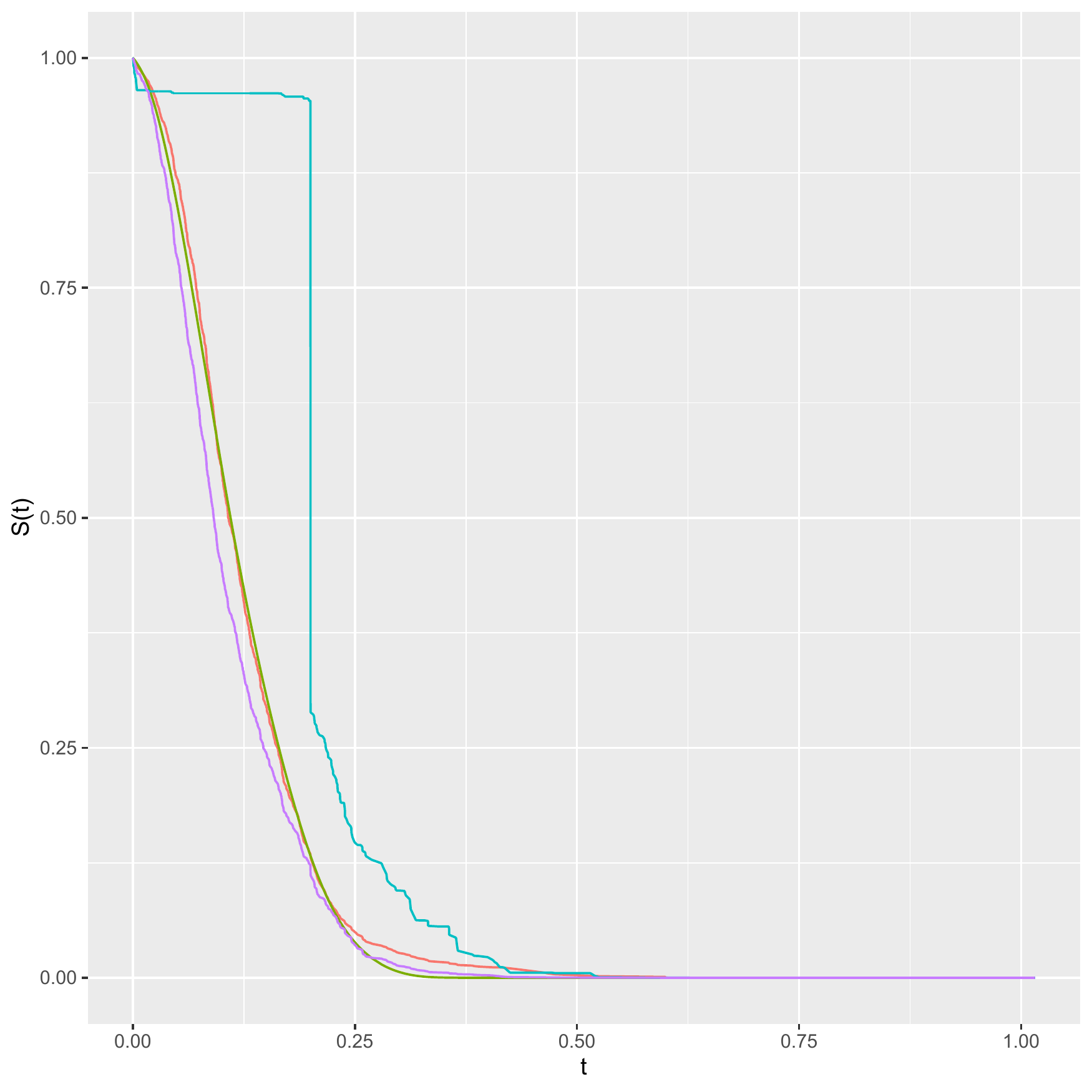}
        \caption{Median-low}
    \end{subfigure}
    \begin{subfigure}[]{0.32\textwidth}
        \centering
    \includegraphics[width=0.95\textwidth]{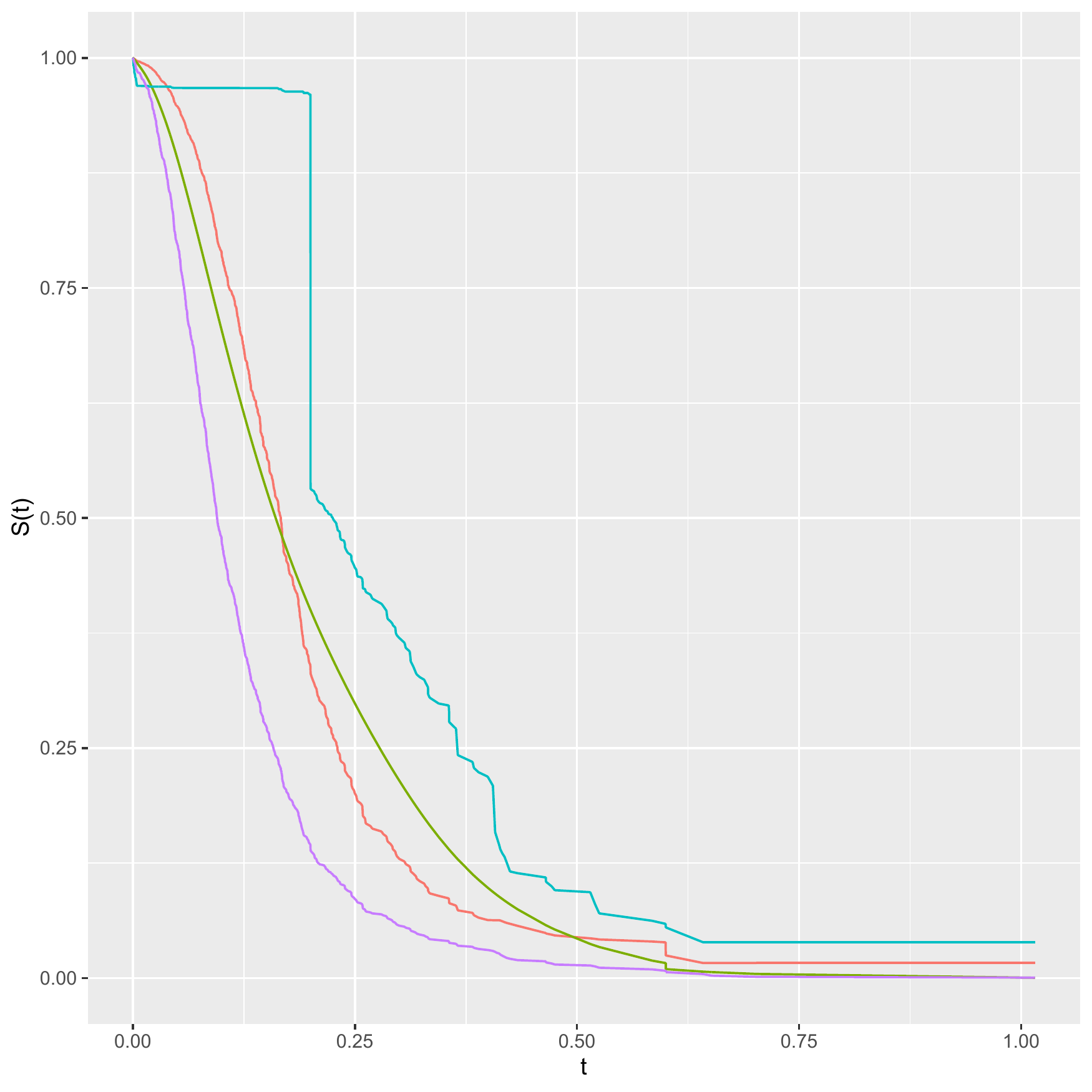}
        \caption{Low}
    \end{subfigure}
       \caption{Survival curves for Model 3 for different groups of subject for $n=200$ (a)-(d) and $n=1000$ (e)-(h).  Red color denotes the proposed DeepHazard, Blue denotes the time-varying Additive Hazards  Method, Green denotes the true Survival curve and Purple denotes the Time-dependent Cox.}\label{fig}
\end{figure}

 For {\bf Model 3}, both for small and large sample, we plot in Figure \ref{fig}, the true and the estimated survival functions by the three different methods. We divide  observations into high, median-high, median-low and low risk according to the risk value $  \sum_{j=1}^4 h_j(Z(t))/4$, i.e., the  mean of the all interval specific risk scores $h_j(Z(t))$.
We observe a strong biased of    the Additive Hazards Model despite a low C-index value. It is often very far from the true survival function.   Figure \ref{fig} part (c) illustrates that $\hat S_{AddHaz}(0.19 \mid Z(0.19))\approx0.875$ while  the true survival function satisfies $ S(0.19 \mid Z(0.19))\approx0.187$. On the other hand $\hat S_{DeepHaz}(t \mid Z(t))$ is a good smooth approximation of the true function. We also observe that larger samples lead to a better Deep Hazard approximation.

Further studies on {\bf Model 3} were done to showcase the impact of the architecture on the learning.  We see that our procedure preforms better both in terms of C-index as well as IMSPE measure of prediction quality. We see that DeepHazards is showcasing IMSPE improvement from $50\%$ to $200\%$.
 
\begin{table}[H] \centering \caption{Model 3 where  each Layer is dense and learning rate is $2e-1$ unless specified differently. Activation function is Relu and $\lambda=1e-5$ with Ridge penalty.}

\begin{tabular}{@{\extracolsep{-5pt}} lccccccccccccccc} 
 \Xhline{.8pt} 
 \multicolumn{16}{c}{Architecture }\\  
 \# of Layers& \multicolumn{2}{c}{One}& &\multicolumn{4}{c}{Two} & & \multicolumn{2}{c}{ Three} && \multicolumn{2}{c}{Four} &   & \multicolumn{1}{c}{ Ten} \\  
 \cline{2-3} \cline{5-8} \cline{10-11} \cline{13-14} \cline{16-16}
Node per layer& \scriptsize [50] &\scriptsize[50]  &&\scriptsize [10]&\scriptsize [10]& \scriptsize[50]&\scriptsize [50] &&\scriptsize[10]& \scriptsize[10] && \scriptsize[10] &\scriptsize [10] &&\scriptsize [10]\\
 Learning rate&    &\scriptsize$2e-2$ &&   &\scriptsize$2e-2$  &   &\scriptsize$2e-2$&&  &\scriptsize$2e-2$ &&    &\scriptsize$2e-2$   && \scriptsize$2e-2$   \\
 \Xhline{.8pt} 
 \rowcolor{lightgray}\multicolumn{16}{c}{ IMSPE $*100$} \\ \Xhline{.8pt} 
Deep Hazard & 0.311  & 0.282 && 0.365& 0.287 &0.369&0.423 &&0.409 & 0.316 &&0.315  &0.326 &&0.529\\
Additive Hazards   & $7.373$   & $7.373$ & & $7.373$  & $7.373$  & $7.373$  & $7.373$ &&  $7.373$  & $7.373$ &&  $7.373$  & $7.373$  && $7.373$  \\
Time-varying  Cox   & 0.967& 0.967 && 0.967 & 0.967 & 0.967 & 0.967 && 0.967 & 0.967 && 0.967 & 0.967 && 0.967   \\
 \Xhline{.8pt} 
 \rowcolor{lightgray}\multicolumn{16}{c}{ C-index} \\ \Xhline{.8pt} 
Deep Hazard   &  0.717& 0.708 && 0.710 & 0.714 &0.723 &0.744  && 0.705  & 0.796 && 0.696  &0.710 && 0.695\\
Additive Hazards   & 0.674 &  0.674 & & 0.674 &  0.674 &  0.674 &  0.674 & & 0.674 &  0.674 & & 0.674 &  0.674 & & 0.674  \\
Time-varying  Cox   &$0.683$ &$0.683$&&$0.683$&$0.683$&$0.683$&$0.683$&&$0.683$&$0.683$&&$0.683$&$0.683$&&$0.683$\\
 \cline{2-3} \cline{5-8} \cline{10-11} \cline{13-14} \cline{16-16}
Oracle   & $0.716$ & $0.716$ && $0.716$ & $0.716$ & $0.716$ & $0.716$ && $0.716$ & $0.716$ && $0.716$ & $0.716$ && $0.716$    \\
\hline \\[-1.8ex] 
\end{tabular}  
\end{table}

 Lastly, we investigated the impact of the activation functions. Setting is that of {\bf Model 3} with Two Layers each comprised of ten (dense) nodes. Learning rate was fixed at $2e-1$.

\begin{table}[H]
\caption{Results within Model 3 across different  activation functions}
\centering
\begin{tabular}{@{\extracolsep{0pt}} lcccccc} 
\Xhline{.8pt}  
 &  {\it Relu} &   {\it Selu} &  {\it Atan} & {\it Tanh}& {\it LogLog} & {\it LeakyRelu}\\
 \Xhline{.8pt} 
 \rowcolor{lightgray}\multicolumn{7}{c}{ IMSPE $*100$} \\ \Xhline{.8pt} 
\cline{2-7}
Deep Hazard& $0.269$ & $0.238$ & $0.308$ & $0.353$ & $0.298$ & $0.399$\\
Additive Hazards    & $7.373$ & $7.373$ & $7.373$ & $7.373$ & $7.373$ & $7.373$    \\
Time-varying  Cox    & $0.967$ & $0.967$ & $0.967$     & $0.967$ & $0.967$ & $0.967$     \\
\Xhline{.8pt} 
 \rowcolor{lightgray}\multicolumn{7}{c}{ C-index} \\ \Xhline{.8pt} 
Deep Hazard   & $0.709$ &  $0.705$ & $0.709$ & $0.688$ &  $0.692$ &  $0.705$\\
Additive Hazards    & $0.674$ & $0.674$ & $0.674$   & $0.674$ & $0.674$ & $0.674$      \\
Time-varying  Cox   & $0.683$ & $0.683$ & $0.683$& $0.683$ & $0.683$ & $0.683$\\
 \cline{2-2}  \cline{3-3}  \cline{4-4}  \cline{5-5}  \cline{6-6}  \cline{7-7} 
Oracle  & $0.716$ & $0.716$ & $0.716$& $0.716$ & $0.716$ & $0.716$\\
\hline \\[-1.8ex] 
\end{tabular} 
\end{table}

\subsection{Impact of a large number of time-varying covariates}

We assume the data is generated according to the following different models:
\begin{itemize}
\item{Model 5: 
\begin{align*}
\lambda(t\mid Z)&=4t^3+\cos(t)[Z_1(t)*Z_2(t)] + | \log(t+1) | Z_1(t)*Z_2(t)
\\
&\qquad +t^3Z_3(t)^2+\frac{1}{1+Z_{20}(t)*Z_1(t)+\sqrt{t}}
\end{align*}
}
\item{Model 6: 
\begin{align*}
\lambda(t\mid Z)
&=4t^3+\cos(t)[Z_1(t)*Z_2(t)]+ |\log(t+1) | Z_3(t)*Z_4(t)+t^3Z_5(t)^2
\\
&+\cos[Z_6(t)*Z_7(t)]+Z_8(t)*Z_9(t)+\frac{1+t^2}{t+1}Z_{10}(t)*Z_{11}(t)
\\
& +Z_{12}(t)^3*Z_{13}(t)^4+\frac{1}{1+Z_{20}(t)*Z_{14}(t)+\sqrt{t}}
\end{align*}
}
\end{itemize}
 where $Z_i(t)$ follows \eqref{eq:Z}.
 
 For {\bf Model 5} all $Z_{0i}$ are drawn from $U(0,20)$ except for $Z_{01}$ that is drawn from $U(5,20)$ and $Z_{020}$,$Z_{019}$ from $U(3,4)$ and $Z_{16}$,$Z_{17}$,$Z_{18}$ from $U(0,1)$. For {\bf Model 6} all $Z_{0i}$ drawn from $U(0,20)$ except for $Z_{01}$ from $U(5,20)$ and $Z_{020}$,$Z_{019}$,$Z_{04}$ from $U(3,4)$ and $Z_{16}$,$Z_{17}$,$Z_{18}$ from $U(0,1)$. 
 
 We considered the following measurement times $0.001,0.2,0.4,0.6$ for {\bf Model 5} and at $0.001,0.1,0.2,0.3$ for {\bf Model 6}. We generate 1000 observation for the training set and for the test set.  The Hyperparameters chosen for our neural network are reported in Table \ref{ti2}. 1000 epochs are used with early stopping rate $1e^{-5}$ and initialization method he Normal is employed. The C-index of each model is presented in Table \ref{result2}.  In these cases we observe strong failure of the additive hazards model with C-index being extremely low, especially for non-linear time interactions. Time-varying Cox approach had difficulties due  to the periodic covariate effects.

\begin{table}[H] \centering \caption{Results of Simulation for Additive Hazards Model, Time-dependent Cox and our method (DeepHazard) for Model 5 and 6.}\label{result2}
\def\arraystretch{1.2}
\begin{tabular}{@{\extracolsep{1pt}} lcccc} 
\Xhline{.8pt} 
 \rowcolor{lightgray}\multicolumn{3}{c}{ C-index} \\ \Xhline{.8pt} 
$n=1000$ & Model 5 & Model 6 \\
\hline
{Deep Hazard}  &\cellcolor{charcoal!60}$0.691$&\cellcolor{charcoal!60}$0.635$ \\
{Additive Hazards }   & $ 0.135$  & $0.423$ \\
Time-varying {Cox}   & $0.677$ & $0.598$\\
\hline \\[-1.8ex] 
\end{tabular} 
\end{table}

\begin{table}[H] \centering \caption{DeepHazard experimental Hyperparameters for Model 5 and 6. }
\begin{tabular}{@{\extracolsep{5pt}} lcc} \label{ti2}{\it Hyperparameter} &  Model 5 & Model 6\\
\hline
Optimizer  & {\it Sgd}  & {\it Adam} \\
Activaction   & {\it Elu}$(0.1)$    & {\it Selu}  \\
N. Dense Layer   & $1$& $1$ \\ 
N. Nodes Layer & 20 & 20 \\
Learning rate & $2e-1$ & $2e-1$  \\
$\lambda$  & $0.56$ & $0.1$ \\
Penalty & Ridge & Ridge \\
Dropout   & $0.2$    & $0.2$ \\
\end{tabular} 
\end{table}

\subsection{Impact of the censoring rate}
We assume the data is again generated according to the  {\bf Model 4},
with $Z_i(t)$ following \eqref{eq:Z}.
 and
 $Z_{0i}\sim U(0,20)$ for $i=1,2,3$. We assume to measure the covariates at the following times $0.001,0.2,0.4,0.6$. We generate our data under different censoring scenario: $10\%,20\%$. We also consider the setting of {\bf Model 5} and {\bf Model 6} with covariates measured at $0.001,0.1,0.2,0.3$  and $0.001,0.1,0.15,0.2$, respectively, each with censoring of $0\%, 15\%,$ and $30\%$.

  We generate 1000 observations for the training set and for the test set.  The Hyperparameters chosen for our neural network are reported in Table \ref{ti3}. 1000 epochs are used with early stopping rate $1e^{-5}$ and he-Normal initialization. The C-index of each model is presented in Table \ref{result3}. The result shows  strong stability  with respect to  censoring.
  
   \definecolor{amber}{rgb}{1.0, 0.75, 0.0}
\definecolor{bazaar}{rgb}{0.6, 0.47, 0.48}
\definecolor{babyblueeyes}{rgb}{0.63, 0.79, 0.95}
\definecolor{bleudefrance}{rgb}{0.19, 0.55, 0.91}
\definecolor{britishracinggreen}{rgb}{0.0, 0.26, 0.15}
\definecolor{cardinal}{rgb}{0.77, 0.12, 0.23}

\begin{table}[H] \centering \caption{C-index for additive Hazards Model, Time-dependent Cox and our method (DeepHazard) under different censoring scenarios.  }\label{result3}
\def\arraystretch{1.2}
\begin{tabular}{@{\extracolsep{1pt}} lccccccccccccc} 
 \Xhline{.8pt} 
 \rowcolor{lightgray}\multicolumn{11}{c}{ C-index} \\ \Xhline{.8pt} 
 & \multicolumn{2}{c}{ Model 4} & & \multicolumn{3}{c}{ Model 5} &&\multicolumn{3}{c}{ Model 6}   \\  
 \cline{2-3} \cline{5-7} \cline{9-11}
Censoring &  $10\%$ &  $20\%$ & & $0\%$&  $15\%$ & $30\%$&  &$0\%$&  $15\%$ &$30\%$\\
\hline
 {DeepHazard} &  \cellcolor{charcoal!60}$0.724$ & \cellcolor{charcoal!60}$0.719$ &  &\cellcolor{charcoal!60}0.682&\cellcolor{charcoal!60}0.678 & \cellcolor{charcoal!60}0.681& &   $\cellcolor{charcoal!60}{0.641}$ &\cellcolor{charcoal!60}${0.632}$ & \cellcolor{charcoal!60}${0.623}$  \\ 
{Additive Hazards }   & $0.532$    & $ 0.625$ &&0.504& 0.501 & 0.413&& 0.506 & 0.498  & 0.417\\
{Time-dependent Cox}    & $0.713$& $0.699$ &&0.676& 0.671 & 0.674&& 0.604 &0.592  & 0.598 \\
 \Xhline{.8pt} 
\end{tabular} 
\end{table}

\begin{table}[H] \centering \caption{DeepHazard experimental Hyperparameters}
\begin{tabular}{@{\extracolsep{1pt}} lccccccccc} \label{ti3}{\it Hyperparameter} & Model 4 & Model 5 &Model 6& \\
\hline
Censoring & $(10\%,  20\%)$ & $(0\%, 15\%,30\%)$ &(0\%,15\%,30\%)\\
Optimizer  &{\it Adam}   & {\it Sgd}   & {\it Sgd}\\
Activation   & {\it Selu}      & {\it Elu}$(0.7)$  & {\it Elu}$(0.5)$\\
N. Dense Layer   & $2,1$  &2,2,3   &2\\ 
N. Nodes Layer & 10,  20  &20& 20\\
Learning rate & $2e-1,3e-3$ &$1e-2,1e-2,1e-1$  & $1e-2$\\
$\lambda$  & $1e-5,1e-4$  &$0.061,0.061,0.05$&$0.061$\\
Penalty & Ridge   & Lasso & Lasso \\
Dropout   & $0.2$      & $0.1/0.15, 0.1/0.15, 0.1/0.15/0.15$ & $0.1/0.15$\\
\hline \\[-1.8ex] 
\end{tabular} 
\end{table}

Further studies on the impact of the censoring and   architecture structure were performed under {\bf Model 2}.
We worked with $1000$ samples in the training and testing phase and report the findings in Table \ref{tab:7}.

\begin{table}[H] \centering \caption{C-index and IMSPE for Deep Hazard, Additive Hazards and Time-dependent Cox model for Model 2. For the Deep Hazard, the dropout rate is $0.2$ and $\lambda=1e-3$ with Ridge penalty and Adam optimizer is used. Architecture, activation function and learning rate is specified in the table. } \label{tab:7}
\begin{tabular}{@{\extracolsep{1pt}} lccccccccc} 
 \Xhline{.8pt} 
Architecture & \multicolumn{2}{c}{10/10, Relu, $2e-2$} & & \multicolumn{2}{c}{ 20/20, Leaky Relu, $2e-3$}    \\  
 \cline{2-3} \cline{5-6}  
  Censoring & 0\%& 10\%    & & 0\% & 10\%& \\
 \Xhline{.8pt} 
 \rowcolor{lightgray}\multicolumn{7}{c}{ IMSPE$*10$} \\ \Xhline{.8pt} 
Deep Hazard   & $0.045$ & $0.051$  & & $0.036$ & $0.048$\\
Additive Hazards      & $ 0.641$ & $0.785$  && $ 0.641$ & $0.785$  \\
Time-dependent Cox    & $0.249$ & $0.340$  &     & $0.249$ & $0.340$    \\
 \Xhline{.8pt} 
 \rowcolor{lightgray}\multicolumn{7}{c}{ C-index} \\ \Xhline{.8pt} 
Deep Hazard     & $0.735$ & $0.746$ & &$0.732$ & $0.743$  \\
Additive Hazards    & $0.592$  & $0.102$     & &$0.592$  & $0.102$   \\
Time-dependent Cox   & $0.718$ & $0.707$ & &$0.718$ & $0.707$\\
\hline \\[-1.8ex] 
\end{tabular} 
\end{table}

\subsection{Effect of shifting the time points at which the covariates are measured}
We assume the data is again generated according to the {\bf Model 6}.
The covariates are assumed to be measured at the following different sets of time points:
 (A) $0.001,0.1,0.15,0.2$; (B) 
  $0.001,0.05,0.08,0.12$; (C) $0.001,0.15,0.2,0.25$; (D) $0.001,0.05,0.08,0.12,0.15,0.2$.
 
We generate 1000 observation for the training set and for the test set.   The Hyperparameters chosen for our neural network are reported in Table \ref{ti4}. 1000 epochs are used with early stopping rate $1e^{-5}$ and initialization method he Normal. The C-index of each Model is presented in Table \ref{result4}. Our methods outperforms the other traditional ones for every sets of time points. Moreover, it is interesting to notice how, while the C-index of Ls and Cox depends on where the covariates are measured, our method presents greater  stability  with respect to the shift. Our C-index is indeed roughly always $0.63$ no matter at which and how many time points the measurements are taken.

\begin{table}[H] \centering \caption{Results of Model 6 for additive Hazards Model, Time-dependent Cox and our method (DeepHazard) for different censoring scenario.}\label{result4}
\begin{tabular}{@{\extracolsep{5pt}} lcccc} 
\Xhline{.8pt} 
 \rowcolor{lightgray}\multicolumn{5}{c}{ C-index} \\ \Xhline{.8pt} 
 & (A) & (B) & (C) & (D) \\
\hline
 {DeepHazard}  & \cellcolor{charcoal!60}$0.633$ & \cellcolor{charcoal!60}$0.630$ & \cellcolor{charcoal!60}$0.633$ & \cellcolor{charcoal!60}$0.632$ \\
 {Additive Hazards }  & $0.506$    & $ 0.572$ & $0.485$ & $0.605$ \\
 {Time-dependent Cox}   & $0.604$& $0.620$ & $0.601$ & $0.619$ \\
\hline \\[-1.8ex] 
\end{tabular} 
\end{table}

\begin{table}[H] \centering \caption{DeepHazard experimental Hyperparameters}
\begin{tabular}{@{\extracolsep{5pt}} lcccc} \label{ti4}
Time points & A & B & C & D \\
\hline
{\it Hyperparameters} &  & & &   \\
Optimizer  &  \it Adam  &\it  Adam  &  \it Adam  &  \it Adam  \\
Activaction   & {\it Elu}$(0.5)$    & {\it Elu}$(0.5)$ & {\it Elu}$(0.5)$ & {\it Elu}$(1.5)$  \\
N. Dense Layer   & $2$& $2$ & $2$ & $2$  \\ 
N. Nodes Layer & 20 & 20 & 20 & 20 \\
Learning rate & $1e-2$ & $1e-2$ & $1e-2$ & $1e-2$ \\
$\lambda$  & $0.061$ & $0.0007$ & $0.08$ & $0.0001$ \\
Penalty & Lasso & Lasso & Lasso & Lasso \\
Dropout   & $0.1/0.15$    & $0.1/0.15$ & $0.1/0.15$ & $0.1/0.15$ \\
\hline \\[-1.8ex] 
\end{tabular} 
\end{table}

\section{Real data experiments} \label{lab:4}
In this section we use our method on three benchmark real datasets.
 
We compare our method with semiparametric additive hazards Model that assumes:
$$
\lambda(t\mid Z)=\lambda_0(t)+\beta Z,
$$
survival random forest, \cite{ishwaran2008random}, as well as Deepsurv of \cite{katzman2018deepsurv}. Deepsurv is a Cox proportional hazards deep neural network that assumes proportionality of the hazard but it doesn't assume linearity of the risk as the standard Cox model:
$$
\lambda(t\mid Z)=\lambda_0(t) \exp\{ h(Z)\}.
$$
We use the R package \emph{Timereg} and the Python package \emph{PySurvival}, respectively to fit the Additive Hazards Model and DeepSurv.

Notice that both DeepSurv and the traditional Cox Model rely on the proportional hazard assumption, under which the ratio of the cumulative hazards between groups is assumed to be constant with time. As a diagnostic, for each of the dataset analyzed, we plot this ratio between groups defined by binary covariates. Not constant line in this type of plot indicates departure from the proportional hazard assumption; see Figure \ref{fig2}.

With slight abuse in notation, as a measure of predictive capability of the models, we report the traditional concordance index, defined as
$$
\mbox{C}_{\mbox{\tiny index}}=\frac{\sum_{i,i'}\mathbbm{1} {(X_i>X_{i'})}\mathbbm{1}{(h(Z_i)<h(Z_{i'}))}\delta_{i'}}{\sum_{i,i'}\mathbbm{1}_{X_i>X_{i'}}\delta_{i'}}.
$$

\subsection{Molecular Taxonomy of Breast Cancer International Consortium dataset (METABRIC)}
The dataset consists of gene expression and clinical features for 1980 breast cancer patients,\citep{curtis2012genomic}. The time variable is time to death and $57.72 \%$ of observations experienced the event. For ease of comparison we use, as training and test set, the same dataset used in \cite{katzman2018deepsurv} where $20\%$ of the data are saved as test set. As covariates 4 gene indicators are used plus hormone treatment indicator, radiotherapy indicator, chemotherapy indicator, ER-positive indicator and age at diagnosis. 

We report in Table \ref{met} the C-index for us, DeepSurv, Semiparametric Additive Hazards Model (LS) and Survival Random Forest. For our Neural Netwok we use  one layer with 40 nodes, Elu activaction function with $\alpha=0.1$, Adam optimizer, learning rate of $0.001$, $\lambda=1e-4$ for Ridge penalty and $0.1$ for Dropout. For DeepSurv we use the hyperparameters reported in their paper. One layer with 41 nodes, Selu activaction function, Adam optimizer, learning rate of $0.010$, $\lambda=10.891$ for Ridge penalty and $0.160$ as Dropout rate. 

 In Table \ref{met}, in parenthesis, we write the result reported by \cite{katzman2018deepsurv} for both DeepSurv and RSF. We plot in Figure \ref{diagnmet} the ratio of the cumulative hazards between four groups defined by the four  patient's clinical features (hormone treatment indicator, radiotherapy indicator, chemotherapy indicator, ER-positive indicator). 

It is clear from the plot how these ratios are not constant with time and therefore how the proportional hazards assumption, on which Deepsurv is based, is violated. From the results, our method indeed outperforms Deepsurv. Moreover it outperforms   random survival forest which we fine-tuned. RSF  C-index, as per tuning, was  very comparable with Deep Surv.

\begin{table}[H] \centering \caption{Results for Metabric dataset  Results in parenthesis are the reported numbers of \cite{katzman2018deepsurv} of the corresponding methods. }\label{met}
\begin{tabular}{@{\extracolsep{5pt}} lc} 
\Xhline{.8pt} 
 \rowcolor{lightgray}\multicolumn{2}{l}{ C-index} \\ \Xhline{.8pt} 
Deep Hazard &\cellcolor{charcoal!60} {\bf 0.664}  \\
Additive Hazards     & 0.645 \\
Deep Surv  &  0.650 (0.654)  \\
RSF & 0.647 (0.619) 
\end{tabular} 
\end{table}

\subsection{Rotterdam and German Breast Cancer Study Group dataset (GBSG)}
The dataset consists of 1546 patients with node-positive breast cancer \citep{schumacher1994randomized}. The time variable is time to death and $90 \%$ experienced the event. Again, as training and test set, we use the same dataset used in \cite{katzman2018deepsurv} where $20\%$ of the data are saved as test set. The testing data consists of 686 patients in a randomized clinical trial that studies the effect of chemotherapy and hormone treatment on survival rate. We report in Table \ref{gbsg} the C-index for us, DeepSurv, Semiparametric Additive Hazards Model (LS) and Survival Random Forest. 

For our Neural Netwok we use  one layer with 40 nodes, Elu activaction function with $\alpha=0.1$, Adam optimizer, learning rate of $0.01$, $\lambda=0.09$ for Ridge penalty and $0.1$ as Dropout rate. For DeepSurv we use the hyperparameters reported in their paper. 1 layer with 8 nodes, Selu activaction function, Adam optimizer, learning rate of $0.154$, $\lambda=6.551$ for Ridge penalty and $0.661$ as Dropout rate.  Moreover, in parenthesis we report the results reported by \cite{katzman2018deepsurv} for both DeepSurv and RSF.  We plot in Figure \ref{diagngbsg} the ratio of the cumulative hazards between 3 groups defined by the 4 binary variables. It is clear from the plot how these ratios are not constant with time and therefore how the proportional hazards assumption, on which Deepsurv is based, is violated. From the results, our method indeed outperforms Deepsurv which is not showing better results than RSF.  Moreover it outperforms RSF as well.

\begin{table}[H] \centering \caption{Results for GBSG dataset} \label{gbsg}
\begin{tabular}{@{\extracolsep{5pt}} lc}
\Xhline{.8pt} 
 \rowcolor{lightgray}\multicolumn{2}{c}{ C-index} \\ \Xhline{.8pt} 
Deep Hazard  & \cellcolor{charcoal!60}{\bf 0.685}   \\
Additive Hazards    & 0.666   \\
Deep Surv    & 0.670 (0.676) \\
RSF   & 0.680 (0.648) 
\end{tabular} 
\end{table}

\subsection{AIDS Clinical Trials Group (ACTG 320)}
The dataset consists of 1151 HIV-infected patients \citep{muche2001applied}. The data come from a double-blind, placebo-controlled trial that compared the three-drugs regime of indinavir, open label zidovudine (ZDV) or stavudine (d4T), and lamivudine (3TC) with the two-drugs regime of zidovudine or stavudine and lamivudine. Patients were eligible for the trial if they had no more than 200 CD4 cells per cubic millimeter and at least three months of prior zidovudine therapy. Randomization was stratified by CD4 cell count at the time of screening.
The primary outcome measured was time to death and $2.26 \%$ of observations has observed death time. 500 observations are saved as test set. 

\begin{figure}
\centering
\begin{subfigure}[]{0.45\textwidth}
\centering
\includegraphics[width=0.95\textwidth]{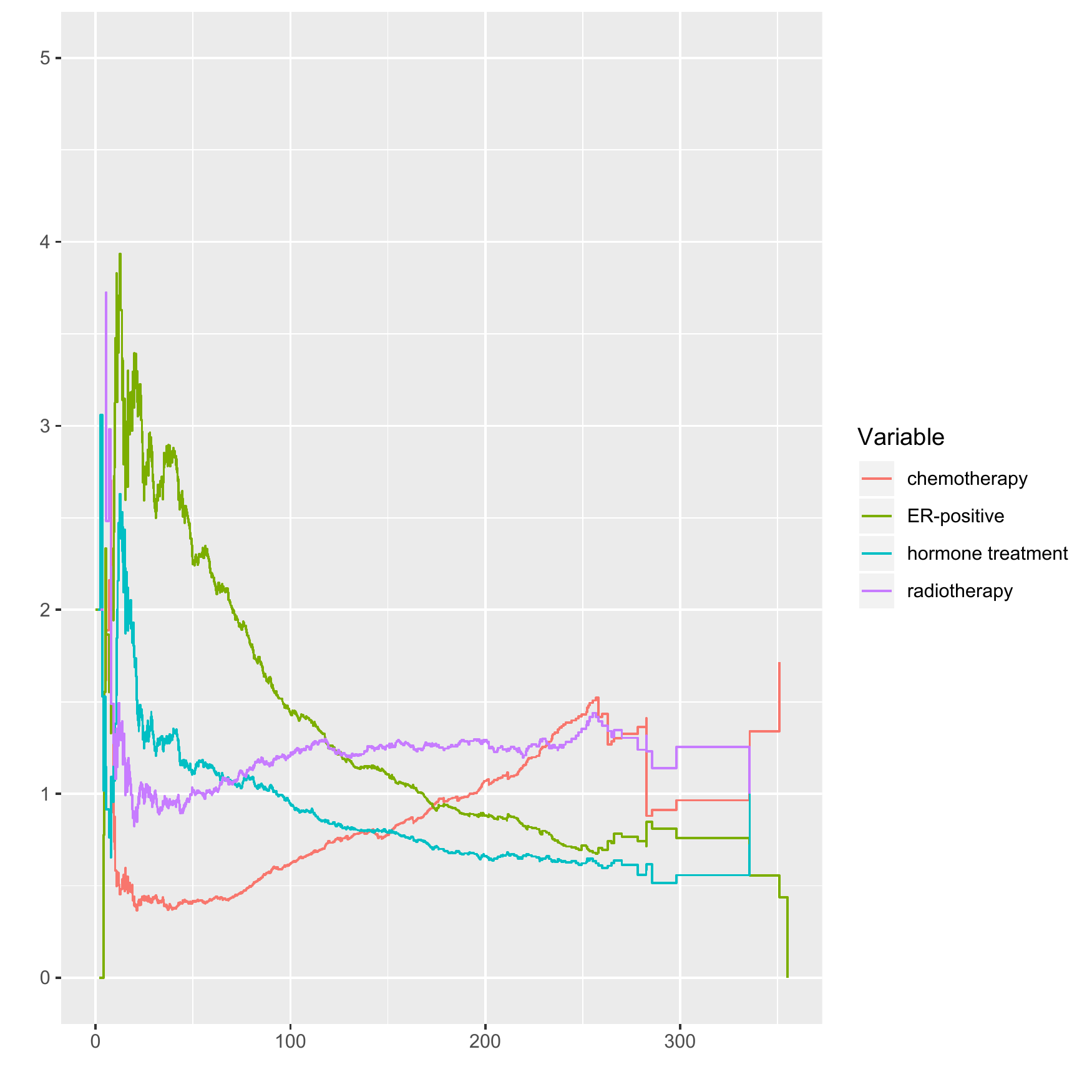}
        \caption{METABRIC on 4 binary variables}\label{diagnmet}
     \end{subfigure}
 \begin{subfigure}[]{0.45\textwidth}
\centering
\includegraphics[width=0.95\textwidth]{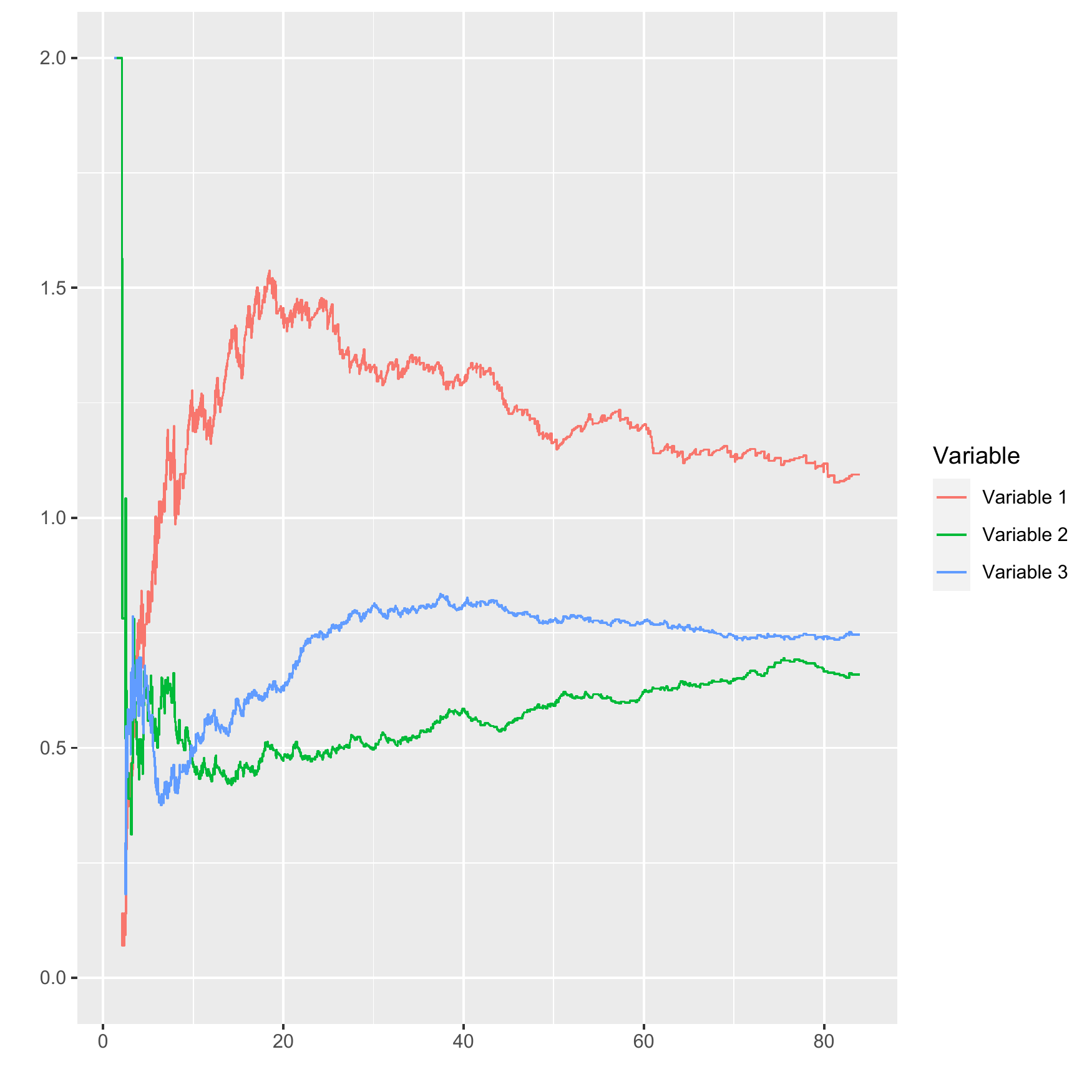}
        \caption{ GBSG on 3 binary variables}\label{diagngbsg}
     \end{subfigure}
\begin{subfigure}{0.45\textwidth}
\centering
\includegraphics[width=0.95\textwidth]{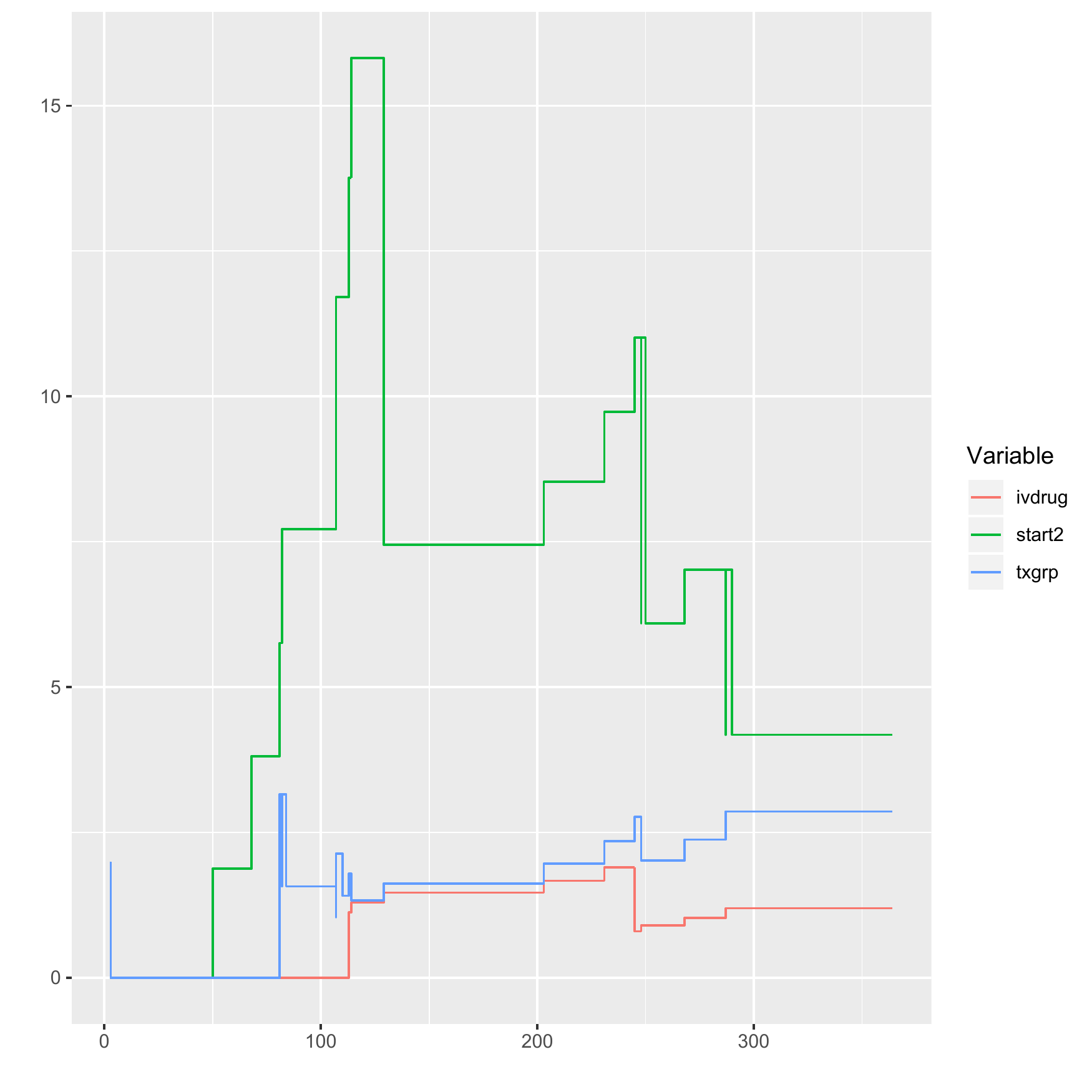}
        \caption{AIDS on 3 binary variables}\label{diagnaids}
     \end{subfigure}
\caption{Proportional hazards diagnostic}\label{fig2}
     \end{figure}

We report in Table \ref{aids} the C-index for DeepHazard, DeepSurv, Semiparametric Additive Hazards Model (LS) and Survival Random Forest. For DeepHazard and DeepSurv we use  2 layers with 50 nodes, Selu activaction function, Adam optimizer, learning rate of $0.1$, $\lambda=2$ with Lasso penalty and $0.2$ as Dropout rate.  We plot in Figure \ref{diagnaids} the ratio of the cumulative hazards between 3 groups defined by 3 binary variables ivdrug, start2 and txgrp, clearly indicating violation of proportionality of the hazards. We observe that our method outperforms DeepSurv and RSF.

\begin{table}[H] \centering \caption{Results for AIDS:ACTG dataset} \label{aids}
\begin{tabular}{@{\extracolsep{5pt}} lc}
\Xhline{.8pt} 
 \rowcolor{lightgray}\multicolumn{2}{c}{ C-index} \\ \Xhline{.8pt}
Deep Hazard  & \cellcolor{charcoal!60}{\bf 0.825}   \\
Additive Hazards      & 0.824   \\
Deep Surv    & 0.773 \\
RSF    & 0.803 \\
\end{tabular} 
\end{table}

     \section{Discussion and possible applications}
Although not extensively exploited in the past due to its complicated interpretation or the lack of methods available, understanding the relationship between survival and time-dependent covariates could be very useful in practice. 

\subsection{Individualized treatments}

It could indeed be a helpful tool for making decisions in the context of dynamic treatment. Let's assume, for example, that, besides some baseline fixed covariates measured at the first visit, $L_0$, at each visit $j$, the doctor has to decide whether to put a patient under treatment, $A_j\in\{0,1\}$, which dose of certain medications to administer, $D_j \in [0,1]$, or whether continue with the same treatment or switch to some alternative, $M_j \in \{1,2,3\}$. The doctor could predict the survival of a new patient under different strategies and pick the one that maximizes patient survival.
For example, at visit 2, considering the history of the covariate of a patient, $Z_{01}=(L_0,A_0.D_0,M_0,A_1,D_1,M_1)$, given two possible different strategies for visit 2, $z_2=(a_2,d_2,m_2)$ and $z_2'=(a'_2,d'_2,m'_2)$, the analysis of the predicted 
$$\hat S(t \mid \{Z_{01},z_2\}), \qquad \mbox{ and } \qquad \hat S(t \mid \{Z_{01},z'_2\})$$
 could help the doctor decides whether to treat the patient with strategy $z_2$ or $z'_2$.
More in general, the same reasoning applies to other varying clinical variables as blood pressure. It could indeed be useful to observe the change in predicted survival under the different hypothetical paths of such covariates. If, for example, the increase of blood pressure appears dangerous for the patient, the doctor could think to introduce medications to keep it stable. 

\subsection{Estimation of treatment effects}

Estimated conditional survival could also be needed as a necessary step  towards obtaining a flexible estimator of some other parameter of interest.
 For example, it is common in the double robust treatment effect estimation literature to employ the use of method that require, as step one, the estimation of baseline quantity or conditional survival distribution. This is in particularly true when AIPW scores are constructed. The augmentation part of the latter indeed usually requires estimation of the conditional distribution of both the censoring and the time to event variable, \cite{zhang2012double,zhao2015doubly,kang2018estimation}.

\subsection{Variable predictive strength}

On the other hand, the estimated conditional survival could be used to estimate other quantities of interest as the expected value of the survival time or $R^2$ measure of explained variation to study the predictive ability of different covariates. The latter is indeed function of the conditional variance of time T and it can be estimated, if an estimator $\hat S(t \mid Z(t))$ is available, using the following formula:
$$
\widehat {\mbox{Var}} \left\{T \mid Z(t)\right\}=\int_0^\tau 2t \hat S(t \mid Z(t)) dt - \left\{\int_0^\tau  \hat S(t \mid Z(t)) dt\right\}^2
$$
Measure of explained variation can be used, for example, to evaluate the clinical importance of prognostic factors, the impact of genetic variants on gene expression on survival phenotypes or they can be applied in variable screening process, \cite{muller2008quantifying,hielscher2010prognostic,kong2019composite}.

  \appendix 
     
     \section{Appendix}
     \subsection{Activation functions} \label{a1}
     \begin{table}[H] \centering \caption{Activation functions} \label{act}
\begin{tabular}{@{\extracolsep{5pt}} ll}
Atan  & $a(x)=\text{atan} (x)$   \\
\\
Elu($\alpha$)   & $a(x)=\begin{cases}x & x>0\\ \alpha (e^x-1) & x \leq 0\end{cases}$   \\
\\
LeakyRelu    & $a(x)=\begin{cases}x & x>0\\ 0.01x & x \leq 0\end{cases}$ \\
\\
LogLog   & $a(x)=1-\exp(-\exp(x))$ \\
\\
Relu & $a(x)=\begin{cases}x & x>0\\ 0 & x \leq 0\end{cases}$\\
\\
Selu & $a(x)=1.0507\begin{cases}x & x>0\\ 1.67326(e^x-1) & x \leq 0\end{cases}$\\
\\
Tanh & $a(x)=\text{tanh} (x)$\\
\end{tabular} 
\end{table}

\subsection{Technical details about $\Bar{h}(t)$} \label{a2}
We explain here why, for each $j$, $\Bar{h}_{j,\theta_j}(t)$ is a step function with jump at censored event time $X^j_i$.
We know that
\begin{equation}
\Bar{h}_{j,\theta_j}(t)=\frac{\sum_{i=1}^{n_j}h_{j,\theta_j}(\tilde Z^j_i)Y^j_i(t)}{\sum_{i=1}^{n_j}Y^j_i(t)}
\end{equation}
and that, by definition,
\begin{equation}
Y^j_i(t)=\begin{cases}
1 & t\leq X^j_i\\
0 & t> X^j_i
\end{cases}
\end{equation}
Therefore, $\Bar{h}_{j,\theta_j}(t)$ represents the mean of ${h}_{j,\theta_j}(\tilde Z^j)$ into the risk set at time $t$. Since the risk set changes only when an individual is censored or dies, $\Bar{h}_{j,\theta_j}(t)$ changes only at censored event time $X^j_i$

 \subsection{Details on the estimation of cumulative hazard}\label{sec:3}
If we break down everything we will have:
\begin{align*}
\lambda(t)&=\sum_{j=1}^{M+1}\lambda_0(t) \mathbbm{1}(t_{j-1}<t\leq t_{j}) +\sum_{j=1}^{M+1}\mathbbm{1}(t_{j-1}<t\leq t_{j}){h}( Z_i(u),u)\\
&=\sum_{j=1}^{M+1}\lambda_0(t) \mathbbm{1}(t_{j-1}<t\leq t_{j}) +\sum_{j=1}^{M+1}\mathbbm{1}(t_{j-1}<t\leq t_{j}){h}_j( \tilde Z^j_i) 
\end{align*}
so if $\lambda^j_0(t)=\lambda_0(t) \mathbbm{1}(t_{j-1}<t\leq t_{j})$, we have:
\begin{align*}
\lambda(t)&=\sum_{j=1}^{M+1}\left[\lambda^j_0(t) +{h}_j( \tilde Z^j_i)\right]\mathbbm{1}(t_{j-1}<t\leq t_{j}) 
\end{align*}
Therefore, if we consider:
\[
dN_i^j (t)= dM_i^j (t)+ \int_{t_j} ^{t}  Y_i^j(u) d \Lambda (u | Z_i(u)), t \in [t_j,t_{j+1})
\]
\[
dN_i^j (t)= dM_i^j (t)+ \int_{t_j} ^{t}  Y_i^j(u) d \Lambda^j (u | \tilde Z^j_i), t \in [t_j,t_{j+1})
\]
we have:
\[
\hat \Lambda^j _0(t) = \begin{cases} 0 & t\leq t_{j}\\
\int_{t_{j}}^t \left[\sum_{i=1}^nY_i^j(u)\right]^{-1} {\sum_{i=1}^n\left (dN_i^j(u)-Y_i^j(u)\hat{h}_j({\tilde  Z}_i^j)\right)}du & t_j<t\leq t_{j+1}\\
\int_{t_{j}}^{t_{j+1}} \left[\sum_{i=1}^nY_i^j(u)\right]^{-1} {\sum_{i=1}^n\left (dN_i^j(u)-Y_i^j(u)\hat{h}_j({\tilde  Z}_i^j)\right)}du & t>t_{j+1}
\end{cases}
\]
and so:
\begin{align*}
\Lambda_0(t)&= \sum_{j=1}^{J-1: t_{J-1}<t<t_{J}}\int_{t_{j-1}}^{t_{j}}\lambda_0(t)+\int_{t_{J-1}}^t\lambda_0(t)=\sum_{j=1}^{J: t_{J-1}<t<t_{J}}\int_{t_{j-1}}^{t_{j}}\lambda^j_0(t)+\int_{t_{J-1}}^t\lambda^J_0(t)\\
\end{align*}
and so:
\[
\hat{\Lambda}_0(t)=\sum_{j=1}^{M+1}\hat{\Lambda}^j_0(t)\\
\]
and therefore:
\begin{align*}
\hat\Lambda_0(t)&=\sum_{j=1}^{M+1}[\int_{t_{j-1}}^{t_{j}}\left[\sum_{i=1}^nY_i^j(u)\right]^{-1} {\sum_{i=1}^n\left (dN_i^j(u)-Y_i^j(u)\hat{h}_j({\tilde  Z}_i^j)\right)}du\mathbbm{1}\{t>t_{j}\}\\&+\int_{t_{j}}^t\left[\sum_{i=1}^nY_i^j(u)\right]^{-1} {\sum_{i=1}^n\left (dN_i^j(u)-Y_i^j(u)\hat{h}_j({\tilde  Z}_i^j)\right)}du\mathbbm{1}\{t_j<t<t_{j+1}\}]\\
&=\sum_{j=1}^{M+1}[\int_{t_{j-1}}^{t_{j}}\left[\sum_{i=1}^nY_i(u)\right]^{-1} {\sum_{i=1}^n\left (dN_i(u)-Y_i(u)\hat{h}(Z(u),u)\right)}du\mathbbm{1}\{t>t_{j}\}\\
&+\int_{t_{j}}^t\left[\sum_{i=1}^nY_i(u)\right]^{-1} {\sum_{i=1}^n\left (dN_i(u)-Y_i(u)\hat{h}(Z(u),u)\right)}du\mathbbm{1}\{t_j<t<t_{j+1}\}]\\
&=\int_0^t\frac{\sum_{i=1}^n\left\{dN_i(u)-Y_i(u)\hat{h}( Z_i(u),u)du\right\}}{\sum_{i=1}^nY_i(u)}
\end{align*}

\bibliographystyle{apalike}
\bibliography{DeepHazard}

\end{document}